\documentclass[10pt,twocolumn,letterpaper]{article}
\usepackage[letterpaper,textwidth=6.875in,textheight=8.875in,top=1in,columnsep=.3125in]{geometry}
\usepackage[numbers,sort&compress]{natbib}
\usepackage[font=small,labelfont=bf]{caption}
\usepackage[dvipsnames]{xcolor}
\usepackage{xspace}
\newcommand{\eg}{\emph{e.g.}\xspace}
\newcommand{\ie}{\emph{i.e.}\xspace}
\date{}

\usepackage[T1]{fontenc}
\usepackage{times}
\usepackage{graphicx}
\usepackage{amsmath,amssymb,bm}
\usepackage{booktabs,multirow,array,tabularx}
\usepackage{microtype}
\usepackage{tikz}
\usetikzlibrary{arrows.meta,positioning,fit,backgrounds,calc}
\usepackage{enumitem}
\usepackage{url}
\newcommand{\method}{GlyphAtlas}
\newcommand{\R}{\mathbb{R}}

\newcommand{\cP}{\mathcal{P}}
\newcommand{\cS}{\mathcal{S}}
\newcommand{\cG}{\mathcal{G}}
\newcommand{\cA}{\mathcal{A}}
\newcommand{\cC}{\mathcal{C}}
\newcommand{\cK}{\mathcal{K}}

\DeclareMathOperator{\Render}{Render}
\definecolor{cvprblue}{rgb}{0.21,0.49,0.74}
\usepackage[pdfusetitle,breaklinks,colorlinks,allcolors=cvprblue]{hyperref}

\usepackage[capitalize]{cleveref}
\crefname{section}{Sec.}{Secs.}
\crefname{table}{Tab.}{Tabs.}
\setlist[itemize]{noitemsep,leftmargin=*,topsep=0pt}
\newcommand\mypara[1]{\par\vspace{1.0mm}\noindent\textbf{#1}}
\let\originalparagraph\paragraph
\RenewDocumentCommand\paragraph{s o m}{\mypara{#3}}

\title{GlyphAtlas: Disentangling Stroke Detail \texorpdfstring{\\}{}from Layout for Scene Text Synthesis}
\author{Luozheng Qin\textsuperscript{1,*} \qquad Changhai Zhou\textsuperscript{2,*} \qquad Yiyi Zhou\textsuperscript{1,\#}\\[3pt]
\normalsize \textsuperscript{1}Xiamen University \qquad \textsuperscript{2}Fudan University\\
\small * Equal contribution \qquad \# Corresponding author}
\hypersetup{pdfauthor={Luozheng Qin, Changhai Zhou, Yiyi Zhou}}
\makeatletter
\newcommand{\preprinttitle}{%
  {\centering\Large\bfseries\@title\par}
  \vspace{6pt}
  {\centering\large\begin{tabular}{c}\@author\end{tabular}\par}}
\makeatother
\begin{document}
\twocolumn[{%
\begin{minipage}{\textwidth}
\preprinttitle
\vspace{9pt}
\centering
\resizebox{0.95\textwidth}{!}{\includegraphics{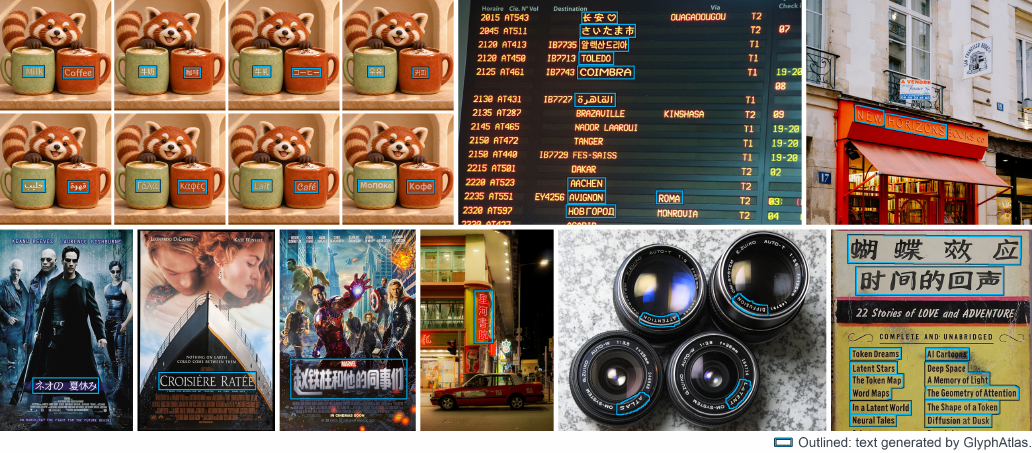}}\par
\captionsetup{type=figure,skip=0pt}
\caption{
    \textbf{\method\ renders stroke-accurate text that blends seamlessly into real scenes.}
    It keeps every text region correct in dense, multi-line layouts,
    harmonizes typography with scene geometry and lighting,
    stays legible when glyphs are small, dense, or structure-complex,
    and generalizes to scripts unseen during training.
}
\label{fig:teaser}
\par\vspace{8pt}
\end{minipage}
}]
\renewenvironment{abstract}
  {{\centering\large\bfseries Abstract\par}\vspace{4pt}\small\noindent\ignorespaces}
  {\par}
\begin{abstract}
Scene text synthesis requires stroke-accurate characters rendered in a specified layout, as well as visual coherence with the surrounding scene.
To achieve this, existing methods commonly pre-render the target text at its intended locations to form a glyph image that jointly specifies layout and glyph structure.
However, this representation ties the resolution of each glyph to its target region, limiting fine-grained stroke guidance for small, densely packed, or structurally complex text.
To tackle this, we propose \method, a scene text synthesis framework that retains the glyph image as a spatial anchor and supplies fine-grained stroke references through a dedicated glyph atlas.
The atlas renders and encodes each character or word in a separate cell, decoupling reference resolution from target-region size.
We further introduce atlas-anchored attention, which lets each text region retrieve stroke detail only from its corresponding atlas cells to refine the coarse glyph structure provided by the glyph image.
This coarse-to-fine conditioning simplifies stroke rendering while preserving explicit layout correspondence.
We optimize \method\ with a mask-weighted flow matching objective and a timestep-weighted text perceptual loss for writing accuracy and background consistency.
Random resolution augmentation of atlas cells supports adjustable atlas budgets at inference.
Extensive experiments demonstrate that \method\ achieves stroke-accurate synthesis of small, dense, and structurally complex text, zero-shot transfer to languages unseen during task-specific training, and leading performance in scenes with many text regions. project page: \url{https://fr0zencrane.github.io/project-glyphatlas/}
\end{abstract}

\section{Introduction}
\label{sec:intro}

Scene text synthesis aims to render accurate, legible text in specified regions of a scene image while keeping it visually coherent with its surroundings~\cite{anytext,fluxtext}, supporting applications such as photo editing~\cite{srnet,diffute} and poster design~\cite{postermaker}.
To achieve this, the conditioning strategy in this area usually pre-renders the target text at its intended locations into a glyph image, which jointly specifies layout and glyph structure for a diffusion model~\cite{glyphcontrol,anytext,anytext2,fluxtext,textflux}.
This representation is widely adopted because it is spatially aligned with the scene while specifying glyph structure: the model knows exactly which glyph to generate and where each glyph belongs.
However, this spatial alignment also ties the resolution of each glyph reference to its target region size, leaving small, dense and structure-complex text with limited pixels to represent its strokes~\cite{textflux}.
The model is therefore misled by the flawed stroke cues in the glyph image, yielding incomplete, merged, or distorted characters.

We argue that such failures arise because the glyph image is overloaded with two roles.
As a spatial anchor, it works well, since it consistently conveys where each glyph belongs and roughly what it looks like regardless of region size.
As a stroke reference, however, it degrades with region size.
Prior work has sought to compensate or replace with glyph embeddings extracted by OCR or other visual text encoders~\cite{anytext2,udifftext,glyphbyt5}, but the remedy has three drawbacks.
It introduces an additional encoder, its recognition-oriented features are trained to be invariant to exactly the stroke-level variation that needs to be rendered, and once injected as global tokens the embeddings lose their correspondence with the region they describe.

In light of the above, we propose \textbf{\method}, which decouples the two roles while keeping them collaborative.
The glyph image is retained solely for layout and coarse glyph structure, while fine-grained stroke guidance is supplied by a dedicated glyph atlas.
Specifically, the proposed glyph atlas renders each character or word in a separate cell at a resolution independent of its target region, so that every text region gains access to a detailed stroke reference even when its scene-aligned glyph occupies only a few pixels.
Crucially, atlas cells are encoded with the diffusion backbone's native VAE rather than a dedicated text encoder.
Benefiting from the VAE's pixel-level reconstruction objective, its features preserve low-level stroke structure and remain sensitive to stroke variation, providing detail-rich guidance without introducing any auxiliary encoder.

To bind the atlas back to the layout, we introduce atlas-anchored attention, where scene and glyph-image tokens retrieve stroke detail only from the atlas cells assigned to the text regions they overlap, while each cell attends to the prompt and its own region.
Local stroke evidence thus refines both the scene features and their glyph condition without ever losing regional correspondence.
The resulting coarse-to-fine conditioning simplifies stroke rendering, as the model no longer has to recover fine structure from an under-resolved condition but only to refine a coarse structure already placed in the scene. %
\method\ is trained with a mask-weighted flow matching objective~\cite{flowmatching,rectifiedflow} and a timestep-weighted text perceptual loss, which together target writing accuracy and background consistency.
Since the atlas grows with the number of characters, we further apply random resolution augmentation to atlas cells during training, which lets the atlas token budget be reduced at inference to accommodate scenes with many text regions.

We evaluate \method\ on standard benchmarks and in the settings that strain glyph-image conditioning.
\method\ achieves the highest sentence accuracy among compared methods on AnyText-benchmark~\cite{anytext}, TextOCR~\cite{textocr}, Total-Text~\cite{totaltext}, and ReCTS~\cite{rects}.
On unseen scripts in STIPLAR~\cite{stellar}, it leads all methods without language-specific fine-tuning by large margins.
Ablations trace these gains to the glyph atlas, which improves sentence accuracy by 9.6 points on small Total-Text regions, 8.1 points on Chinese text, and over 20 points on Korean and Arabic over an identically trained glyph-image-only model.
Our main contributions are as follows:
\begin{itemize}
    \item We identify that glyph-image conditioning conflates two roles: it remains a reliable spatial anchor, yet degrades as a stroke reference once regions shrink.
    \item We propose \method, which decouples yet coordinates the two roles: the glyph atlas supplies region-independent stroke references, and atlas-anchored attention binds them back to the regions laid out by the glyph image.
    \item \method\ achieves state-of-the-art performance on four scene text benchmarks, and ablations attribute the gains on small, dense, structurally complex, and unseen-script text to the glyph atlas together with atlas-anchored attention.
\end{itemize}

\section{Related Work}
\subsection{Text-to-Image Diffusion Models}
Diffusion models~\cite{ddpm,ldm} are the dominant paradigm for visual generation, evolving from UNet-based latent diffusion~\cite{ldm,sdxl} to diffusion transformers (DiTs)~\cite{dit} trained with flow matching~\cite{flowmatching,rectifiedflow}, as represented by SD3~\cite{sd3} and FLUX~\cite{flux2024}.
Conditioned on a text prompt alone, these models offer weak control over spatial layout, and visual text exposes this most clearly, since a word must appear at a specific place with a specific spelling~\cite{characteraware}.
The common remedy is an additional image condition, injected through an auxiliary branch in UNets~\cite{controlnet} or concatenated with the noisy latent into a unified sequence in DiTs~\cite{ominicontrol,icedit,fluxkontext}.
Scene text synthesis is an instance of this setup in which the condition is usually a pre-rendered glyph image, and we detail in \cref{subsec:scene} how this condition is built and where it falls short.

\subsection{Scene Text Synthesis}
\label{subsec:scene}
Scene text synthesis covers scene text generation and editing, both of which render legible text in specified regions of a scene image~\cite{anytext,fluxtext}.
Early methods composite rendered text onto natural images~\cite{synthtext} or edit existing text with GANs~\cite{srnet}.
Diffusion models have since become dominant, first as UNet-based inpainting models with auxiliary glyph conditions~\cite{diffste,diffute,textdiffuser,glyphcontrol,anytext} and recently as fine-tuned DiTs~\cite{fluxtext,textflux}.
Across these models, the prevailing condition is a glyph image: the target text pre-rendered in a canonical font at its target locations on a scene-sized canvas~\cite{glyphcontrol,anytext,fluxtext,textflux}.
Being pixel-aligned with the scene, one glyph image serves as both spatial anchor and stroke reference, but each glyph then receives only as many pixels as its region allows, so small, dense, and structure-complex text remains a commonly failure~\cite{postermaker,textflux}.
A region-independent reference lifts this limit but gives up the alignment, which existing work restores along two routes.
The first injects glyph embeddings from OCR recognizers or glyph-aligned text encoders~\cite{anytext2,udifftext,glyphbyt5,postermaker}.
This requires an additional encoder whose character set bounds its coverage~\cite{anytext2}, yields recognition features invariant to the very stroke variation to be rendered, and, unless routed region-wise~\cite{glyphbyt5}, detaches each embedding from the region it describes~\cite{anytext2}.
The second, closer to ours, renders each text item as a separate glyph patch and binds it to its region through positional encodings~\cite{easytext,glyphanchor} or attention masks~\cite{glyphprinter}.
Developed mainly for text-to-image rendering, these methods replace the glyph image rather than complement it, so no coarse structure is placed in the scene to be refined, and text is missed or corrupted once the binding becomes ambiguous~\cite{easytext,glyphanchor}.
\method\ takes neither route exclusively: it keeps the glyph image as spatial anchor and coarse structure, and adds a glyph atlas of VAE-encoded cells, bound region-wise by atlas-anchored attention, solely to refine it.

\begin{figure*}[t]
\centering
\includegraphics[width=.95\textwidth]{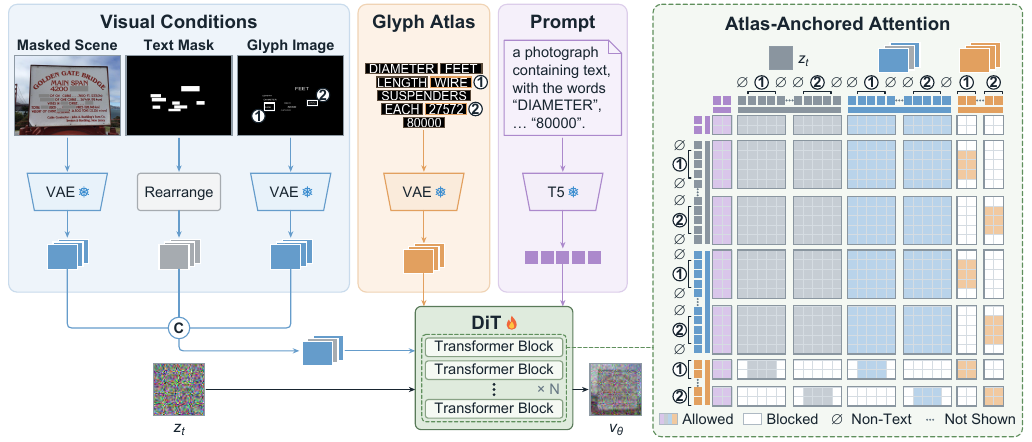}
\vspace{-4pt}
\caption{
Overview of \method.
\method\ conditions a DiT on two complementary glyph references:
(1) the glyph image, which is spatially aligned with the scene and anchors the layout and coarse glyph structure of every text region;
(2) the glyph atlas, which renders each character or word in a separate cell at a resolution independent of its target region and supplies fine-grained stroke references.
Atlas-anchored attention binds the two through an attention mask, so that each text region retrieves stroke detail only from its own atlas cells.
}
\label{fig:overview}
\end{figure*}

\section{Method}
\label{sec:method}

Fig.~\ref{fig:overview} presents an overview of \method.
It conditions a diffusion transformer on two complementary glyph references: the glyph image, which anchors the layout and coarse glyph structure, and the proposed glyph atlas, which supplies fine-grained stroke references and is bound to the layout by atlas-anchored attention.
In the following, we first present the task, the flow matching formulation, and the backbone in Sec.~\ref{sec:prelim}, then introduce the glyph atlas and atlas-anchored attention in Sec.~\ref{sec:atlas}, and finally describe the training objectives and atlas budgeting in Sec.~\ref{sec:train}.

\subsection{Preliminaries}
\label{sec:prelim}

\paragraph{Problem formulation.}
Given a scene image $I$, a prompt $p$ describing the scene, and $K$ text regions $\{(y_r,R_r)\}_{r=1}^{K}$, where $y_r$ is the string to be written and $R_r$ is its target region, scene text synthesis produces an image $\hat I$ in which every $y_r$ is rendered inside $R_r$ while the rest of $I$ is preserved.
This formulation covers both scene text generation and editing, which differ only in whether $R_r$ already contains text.
Existing methods~\cite{anytext,fluxtext,textflux} cast the task as conditional inpainting,
\begin{equation}
\hat I=\mathcal F_\theta\big(I_{\rm mask},\,M,\,G,\,p\big),
\label{eq:task}
\end{equation}
where $\mathcal F_\theta$ is a generative model and the three visual conditions are derived from the inputs and spatially aligned with $I$.
The text mask $M$ is binary and equals one inside $\bigcup_r R_r$, specifying where to write.
The masked scene image $I_{\rm mask}=(1-M)\odot I$ exposes the background to be preserved, with $\odot$ the element-wise product.
The glyph image $G$ renders each $y_r$ inside $R_r$ in a uniform font on a canvas of the same size as $I$, specifying what to write.
Notably, $G$ is the only condition that carries glyph structure, and since it is drawn on a scene-sized canvas, the number of pixels a glyph receives in $G$ is dictated by the size of $R_r$ rather than by the complexity of its strokes.

\paragraph{Flow matching.}
We instantiate $\mathcal F_\theta$ as a flow matching model~\cite{flowmatching,rectifiedflow,sd3}, which, like a diffusion model, generates an image by gradually turning noise into data, but does so along a straight path with a continuous time variable.
Images are first processed by a VAE with encoder $\mathcal E$ and decoder $\mathcal D$~\cite{ldm}, and generation takes place in the latent space.
Given the latent $z_0=\mathcal E(I)$ of a clean image and noise $\epsilon\sim\mathcal N(\mathbf 0,\mathbf I)$, the noisy latent at timestep $t$ is the linear interpolation
\begin{equation}
z_t=(1-t)\,z_0+t\,\epsilon,\qquad t\in[0,1],
\label{eq:path}
\end{equation}
which moves from the clean latent at $t=0$ to pure noise at $t=1$.
Its time derivative, called the velocity, is the constant $v^\star=\mathrm{d}z_t/\mathrm{d}t=\epsilon-z_0$.
A network $v_\theta(z_t,t,c)$ is trained to regress this velocity under $c$, %
which collects the conditions in Eq.~\eqref{eq:task}, with the conditional flow matching loss
\begin{equation}
\mathcal L_{\rm FM}=\mathbb E_{t,\,z_0,\,\epsilon}\,\big\|v_\theta(z_t,t,c)-v^\star\big\|_2^2 .
\label{eq:fm}
\end{equation}
At inference, sampling starts from $z_1=\epsilon$ and integrates $\mathrm{d}z_t/\mathrm{d}t=v_\theta$ from $t=1$ to $t=0$ with an Euler solver over a fixed number of steps, after which $\mathcal D$ maps the result back to pixel space.

\paragraph{Backbone.}
We implement $v_\theta$ in Eq.~\eqref{eq:fm} with FLUX.1 Fill~\cite{flux2024}, an MM-DiT~\cite{sd3} that processes prompt tokens and image tokens as one unified sequence with joint attention in every block.
Its inputs are prepared in three steps.
First, the prompt $p$ is encoded by the frozen T5~\cite{t5} encoder into prompt tokens, while its pooled embedding from the frozen CLIP~\cite{clip} text encoder modulates the timestep embedding, and the visual conditions are brought to the latent resolution, where $I_{\rm mask}$ and $G$ are encoded by $\mathcal E$ and $M$ is rearranged by space-to-depth.
Second, following in-context conditioning~\cite{ominicontrol,fluxkontext}, two image inputs are formed by channel-wise concatenation: the scene stream concatenates $z_t$, $\mathcal E(I_{\rm mask})$, and $M$, and the glyph-image stream concatenates $\mathcal E(G)$ with the same $\mathcal E(I_{\rm mask})$ and $M$.
Third, each stream is patchified and projected to the model width, yielding scene tokens and glyph-image tokens that lie on the same token grid, so that the two tokens at one grid location describe the same patch of the scene.
Prompt, scene, and glyph-image tokens then interact through joint attention in every block, and after the last block the velocity $v_\theta$ is read from the scene tokens and unpatchified to the latent shape.

\subsection{Glyph Atlas and Atlas-Anchored Attention}
\label{sec:atlas}

\method\ retains the glyph image $G$ in Eq.~\eqref{eq:task}, which anchors the layout and provides a spatially aligned coarse glyph reference.
Fine-grained stroke references are supplied by the proposed glyph atlas, a set of tokens appended to the token sequence of the backbone, and bound to the text regions laid out by $G$ through atlas-anchored attention.
We describe the two components in turn, followed by the coarse-to-fine conditioning they jointly realize.

\begin{figure}[t]
\centering
\includegraphics[width=.95\columnwidth]{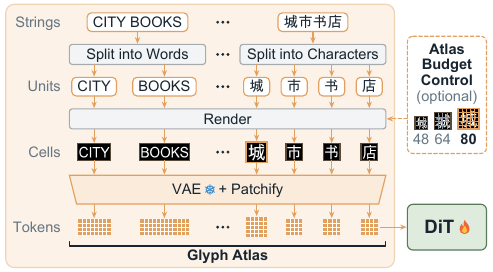}
\vspace{-4pt}
\caption{
Demonstration of the glyph atlas construction.
The height of atlas cell can be adjusted to control the atlas token budget.
}

\label{fig:atlas}
\end{figure}

\paragraph{Glyph atlas.}
The glyph atlas renders the target strings independently of the scene and encodes them into tokens, so that the resolution of a glyph reference no longer depends on the size of its target region.
As shown in Fig.~\ref{fig:atlas}, it is built in three steps.
First, each string $y_r$ is split into an ordered sequence of text units $u_{r1},u_{r2},\dots$, indexed by $j$ within region $r$, where a unit is the smallest span of text read as a whole: a character for scripts whose characters are read individually, such as Chinese and Korean, and a word for alphabetic scripts, such as Latin and Arabic, whose letter spacing and joining forms exist only within a word.
Second, each unit is rendered in a uniform font into its own cell $B_{rj}=\Render(u_{rj};h)$, square for a character and of natural width for a word, at a cell height $h$ that is set freely rather than dictated by $R_r$.
Third, each cell is encoded by the frozen VAE encoder $\mathcal E$ and patchified in the same way as the scene latent,
\begin{equation}
\mathbf a_{rj}=\operatorname{Patchify}\big(\mathcal E(B_{rj})\big)\in\R^{n_{rj}\times d},
\label{eq:atlas}
\end{equation}
where the number of tokens $n_{rj}$ scales with the area of the cell and $d$ is the dimension of a patch.
A cell therefore yields the same tokens whether its glyph occupies a few pixels or a large area in the scene.
We encode cells with the VAE rather than a recognition encoder because the VAE is trained to reconstruct exactly the stroke variation that recognition features are trained to be invariant to, and because its latents already lie in the input space of the DiT.
Accordingly, atlas tokens are projected to the model width by a linear layer initialized from the weights of the DiT input projection that act on the $z_t$ channels.
They receive rotary position embeddings~\cite{roformer} on a separate canvas, one row per region with cells in reading order, together with a learned embedding of the unit index $j$.
Since neither embedding carries scene coordinates, the link between a cell and the scene is established solely by the atlas-anchored attention described next.

\paragraph{Atlas-anchored attention.}
The atlas specifies in detail what each glyph looks like but not where it goes, whereas the glyph image specifies where every glyph goes but only coarsely what it looks like.
Atlas-anchored attention makes the two cooperate by restricting which tokens may attend to which, so that each text region retrieves stroke detail only from its own cells.
To be specific, let $\cP$, $\cS$, and $\cG$ be the index sets of prompt, scene, and glyph-image tokens, $s_i\in\cS$ and $g_i\in\cG$ the two tokens at grid location $i$, and $\cC=\cP\cup\cS\cup\cG$ the base context of the backbone.
Let $\cA_r$ be the atlas tokens of region $r$, \ie, the rows of all $\mathbf a_{rj}$ in Eq.~\eqref{eq:atlas}, and $\cA=\bigcup_{r=1}^{K}\cA_r$ the atlas.
Let $\mathcal O(i)\subseteq\{1,\dots,K\}$ be the set of text regions whose mask overlaps the patch at grid location $i$, which is empty for background locations, and let $\cS_r=\{s_i: r\in\mathcal O(i)\}$ and $\cG_r=\{g_i: r\in\mathcal O(i)\}$ be the scene and glyph-image tokens of region $r$.
Each query $q$ then attends only to the key set
\begin{equation}
\cK(q)=
\begin{cases}
\cC, & q\in\cP,\\[3pt]
\cC\;\cup\;\displaystyle\bigcup_{r\in\mathcal O(i)}\cA_r, & q\in\{s_i,g_i\},\\[3pt]
\cP\cup\cS_r\cup\cG_r\cup\cA_r, & q\in\cA_r.
\end{cases}
\label{eq:binding}
\end{equation}
The three cases read as follows.
Prompt tokens attend only to the base context, so the prompt pathway keeps its pretrained attention pattern.
Scene and glyph-image tokens keep the full base context and additionally attend to the cells of every region they overlap, and attend to no cell at background locations.
The tokens of each cell attend to the prompt, the cells of their own region, and only the scene and glyph-image tokens of that region, so that each reference adapts to the scale and style of the region it serves without being exposed to any other.
The rule is realized as an attention mask over the shared query, key, and value projections of the backbone and introduces no new attention parameters, which fully preserves the pretrained knowledge of the backbone.

\paragraph{Coarse-to-fine conditioning.}
Eq.~\eqref{eq:binding} turns the glyph image into the spatial index of the atlas.
A scene token inside region $r$ sees a coarse rendering of its glyph at the same grid location in the glyph-image tokens and, through $\cA_r$, a high-resolution rendering of the same characters, so its task reduces to refining strokes that are already roughly in place rather than recovering them from an under-resolved condition (Fig.~\ref{fig:overview}, right).
Since glyph-image tokens also attend to the atlas, the coarse condition itself is sharpened before it guides the scene.
Meanwhile, restricting each cell to its own region keeps layout correspondence explicit and prevents the cells of one region from interfering with another, which matters when many regions are present.
Unlike masks that bind a glyph reference only to the generated latent~\cite{glyphprinter}, atlas-anchored attention binds each cell both to its region in the scene and to the coarse rendering of that region in $G$.

\subsection{Optimization and Atlas Budgeting}
\label{sec:train}

Training follows the masked reconstruction setting common in diffusion-based scene text synthesis~\cite{diffute,anytext,fluxtext}, where the target is the scene image $I$ itself, so $z_0$ in Eq.~\eqref{eq:path} is its latent, and the model learns to re-synthesize the text inside $M$ from $I_{\rm mask}$, $M$, $G$, $\cA$, and $p$.
All DiT weights are updated together with the atlas projection and embeddings, while the VAE and the text encoders stay frozen.
Two objectives steer optimization toward legible text, and a resolution augmentation makes the atlas token budget adjustable at inference.

\paragraph{Mask-weighted flow matching.}
We draw $t=\operatorname{sigmoid}(\eta)$ with $\eta\sim\mathcal N(0,1)$, \ie, the logit-normal timestep sampling of SD3~\cite{sd3}, and form $z_t$ and $v^\star$ as in Eq.~\eqref{eq:path}.
Let $m$ be the text mask $M$ downsampled to the token grid and broadcast over channels.
The flow matching loss in Eq.~\eqref{eq:fm} is weighted by the mask,
\begin{equation}
\mathcal L_{\rm flow}=\operatorname{mean}\big[(1+m)\odot(v_\theta-v^\star)^2\big],
\label{eq:flow}
\end{equation}
where the mean is taken over all latent elements.
Text regions thus receive twice the weight of the background, which emphasizes text accuracy without discarding the background term that preserves scene consistency.

\paragraph{Timestep-weighted text perceptual loss.}
Flow matching supervises latents and is agnostic to legibility, so we add a pixel-space loss on the text.
Since $z_0=\epsilon-v^\star$ by Eq.~\eqref{eq:path}, the prediction $v_\theta$ yields a one-step estimate $\hat z_0=\epsilon-v_\theta$ of the clean latent, which we decode into $\tilde I=\mathcal D(\hat z_0)$ with the frozen VAE decoder.
We crop every text region from $\tilde I$ and $I$ and feed the crops to a frozen text recognizer~\cite{ppocrv3}.
Following AnyText~\cite{anytext}, the text perceptual loss $\mathcal L_{\rm OCR}$ is the mean squared difference between the recognizer features of corresponding crops, averaged over valid crops and set to zero when none exists.
The full objective is
\begin{equation}
\mathcal L=\mathcal L_{\rm flow}+2t\,\alpha\,\mathcal L_{\rm OCR},
\label{eq:loss}
\end{equation}
with $\alpha=0.25$.
Since $\mathbb E[t]=1/2$ under logit-normal sampling, the factor $2t$ has unit expectation and $\alpha$ keeps its meaning as the average weight, while supervision shifts toward high-noise inputs, where the structure of the glyphs is decided.
Gradients flow through the frozen decoder and recognizer into the DiT.

\paragraph{Atlas Budgeting.}
The atlas grows linearly with the number of units, and each cell costs tokens in proportion to its area.
Since cell resolution is decoupled from region size, it trades stroke detail against sequence length, which lets scenes with many text regions fit a fixed token budget.
To make the model robust to this trade-off, we add a second training stage after training at a cell height of 64 pixels, in which each non-punctuation unit is re-rendered at 48, 64, or 80 pixels with probabilities 0.2, 0.7, and 0.1, while punctuation stays at 64 pixels.
Consequently, during inference, the user can either set one cell height for all units or specifies a total token budget, which a heuristic allocator distributes across cells, giving smaller scene glyphs higher resolution. As for the sampling procedure, it follows Sec.~\ref{sec:prelim} with $I_{\rm mask}$, $M$, $G$, and $\cA$ held fixed across steps.

\begin{table*}[t]
\centering\small\setlength{\tabcolsep}{4pt}
\caption{
Scene text generation on AnyText-benchmark~\cite{anytext}.
Best results are in \textbf{bold}.}
\vspace{-4pt}
\label{tab:anytext}
\begin{tabular}{lrrrrrrrr}\toprule
\multicolumn{1}{c}{\multirow{2}{*}[-0.45ex]{Method}} & \multicolumn{4}{c}{English} & \multicolumn{4}{c}{Chinese}\\
\cmidrule(lr){2-5}\cmidrule(lr){6-9}
 & ACC $\uparrow$ & NED $\uparrow$ & FID $\downarrow$ & LPIPS $\downarrow$ & ACC $\uparrow$ & NED $\uparrow$ & FID $\downarrow$ & LPIPS $\downarrow$ \\
\midrule
DiffSTE~\cite{diffste}           & 45.23 & 0.7814 & 52.74 & 0.1816 & 3.63  & 0.1226 & 57.49 & 0.1276 \\
TextDiffuser~\cite{textdiffuser} & 51.76 & 0.7618 & 29.76 & 0.1564 & 5.59  & 0.1218 & 34.19 & 0.1252 \\
DiffUTE~\cite{diffute}           & 40.54 & 0.7005 & 25.35 & 0.1640 & 29.78 & 0.5744 & 29.08 & 0.1745 \\
AnyText~\cite{anytext}           & 68.43 & 0.8588 & 21.59 & 0.1106 & 64.76 & 0.8210 & 20.01 & 0.0943 \\
TextCtrl~\cite{textctrl} & 58.53 & 0.8146 & 35.73 & 0.1978 & 35.80 & 0.6084 & 49.79 & 0.2298 \\
DreamText~\cite{dreamtext} & 64.95 & 0.8466 & 28.37 & 0.1314 & 59.03 & 0.8270 & 58.22 & 0.0963 \\
AnyText2~\cite{anytext2} & 66.79 & 0.8610 & 32.59 & 0.1762 & 63.29 & 0.8075 & 29.23 & 0.1383 \\
TextSSR~\cite{textssr} & 58.64 & 0.7664 & 24.23 & 0.1245 & 50.33 & 0.7018 & 29.41 & 0.1063 \\
TextFlux~\cite{textflux}         & 74.81 & 0.8789 & 13.15 & 0.0845 & 54.73 & 0.7705 & 13.02 & 0.0577 \\
FLUX-Text~\cite{fluxtext}        & 84.19 & 0.9400 & 13.85 & \textbf{0.0729} & 71.32 & 0.8510 & 13.68 & 0.0541 \\
\midrule
\method & \textbf{88.68} & \textbf{0.9555} & \textbf{11.99} & 0.0759 & \textbf{77.92} & \textbf{0.8806} & \textbf{11.50} & \textbf{0.0534} \\
\bottomrule
\end{tabular}
\end{table*}

\newsavebox{\sideAcap}\newsavebox{\sideAtab}\newsavebox{\sideBcap}\newsavebox{\sideBtab}
\newlength{\sideH}\newlength{\sideHB}
\begin{table*}[t]
\footnotesize\setlength{\tabcolsep}{2pt}%
\sbox{\sideAtab}{\begin{tabular}{lrrrrrrrrr}\toprule
\multicolumn{1}{c}{\multirow{2}{*}[-0.45ex]{Method}} & \multicolumn{3}{c}{TextOCR} & \multicolumn{3}{c}{Total-Text} & \multicolumn{3}{c}{ReCTS}\\
\cmidrule(lr){2-4}\cmidrule(lr){5-7}\cmidrule(lr){8-10}
 & ACC$\uparrow$ & NED$\uparrow$ & FID$\downarrow$ & ACC$\uparrow$ & NED$\uparrow$ & FID$\downarrow$ & ACC$\uparrow$ & NED$\uparrow$ & FID$\downarrow$ \\
\midrule
AnyText~\cite{anytext}     &  8.12 & 0.1738 &  5.48 & 24.74 & 0.5139 & 31.45 & 34.81 & 0.5071 & 22.19 \\
DreamText$^\dagger$~\cite{dreamtext} & 7.47 & 0.1725 & 18.92 & 27.09 & 0.5217 & 52.27 & 48.20 & 0.7017 & 54.59 \\
AnyText2~\cite{anytext2}   &  9.54 & 0.2084 & 13.20 & 27.40 & 0.5422 & 73.71 & 35.86 & 0.5091 & 39.64 \\
FLUX-Text~\cite{fluxtext}  & 24.11 & 0.3614 &  2.60 & 64.06 & 0.8111 & 16.87 & 48.35 & 0.6477 & 14.58 \\
TextFlux~\cite{textflux}   & 23.62 & 0.3653 & \textbf{2.10} & 62.29 & 0.7829 & \textbf{13.58} & 43.02 & 0.5930 & \textbf{11.97} \\
TextSSR~\cite{textssr} & 27.74 & 0.4499 & 2.71 & 47.06 & 0.6835 & 26.60 & 50.23 & 0.6876 & 20.96 \\
\midrule
\method & \textbf{33.15} & \textbf{0.5046} & 2.81 & \textbf{73.17} & \textbf{0.8647} & 16.05 & \textbf{69.27} & \textbf{0.8259} & 13.27 \\
\bottomrule
\end{tabular}}%
\sbox{\sideBtab}{\begin{tabular}{lrrrrrr}\toprule
\multicolumn{1}{c}{\multirow{2}{*}[-0.45ex]{Method}} & \multicolumn{2}{c}{Japanese} & \multicolumn{2}{c}{Korean} & \multicolumn{2}{c}{Arabic}\\
\cmidrule(lr){2-3}\cmidrule(lr){4-5}\cmidrule(lr){6-7}
 & ACC$\uparrow$ & FID$\downarrow$ & ACC$\uparrow$ & FID$\downarrow$ & ACC$\uparrow$ & FID$\downarrow$ \\
\midrule
AnyText2~\cite{anytext2}  & 47.79 & 105.62 & 29.82 & 63.52 &  0.10 & 99.32 \\
FLUX-Text~\cite{fluxtext} & 46.49 &  80.73 & 20.83 & 44.39 &  2.65 & 77.30 \\
TextFlux~\cite{textflux}  & 46.23 &  \textbf{67.43} & 33.86 & \textbf{24.01} &  6.42 & \textbf{51.96} \\
STELLAR$^\ddagger$~\cite{stellar} & 43.38 & 73.73 & \textbf{80.47} & 34.47 & \textbf{55.86} & 62.84 \\
TextSSR~\cite{textssr} & 40.26 & 102.86 & 19.14 & 68.11 & 0.20 & 92.54 \\
\midrule
\method-Bi                & \textbf{67.53} & 84.09 & 66.96 & 38.99 & 39.55 & 68.10 \\
\bottomrule
\end{tabular}}%
\sbox{\sideAcap}{\begin{minipage}[t]{\wd\sideAtab}
\caption{
Scene text editing on three real-scene benchmarks, TextOCR~\cite{textocr}, Total-Text~\cite{totaltext} and ReCTS~\cite{rects}.}
\vspace{-4pt}
\label{tab:scene}
\end{minipage}}%
\sbox{\sideBcap}{\begin{minipage}[t]{\wd\sideBtab}
\caption{
Zero-shot multilingual scene text editing on STIPLAR~\cite{stellar}.
$\ddagger$: separate model tuned per language.
}
\label{tab:stiplar}
\end{minipage}}%
\sideH=\dimexpr\ht\sideAcap+\dp\sideAcap+\ht\sideAtab+\dp\sideAtab\relax
\sideHB=\dimexpr\ht\sideBcap+\dp\sideBcap+\ht\sideBtab+\dp\sideBtab\relax
\typeout{SIDEBYSIDE A=\the\sideH\space B=\the\sideHB}%
\noindent\begin{minipage}[t]{\wd\sideAtab}\offinterlineskip
\usebox{\sideAcap}\par
\ifdim\sideHB>\sideH\vskip\dimexpr\sideHB-\sideH\relax\fi
\usebox{\sideAtab}
\end{minipage}\hfill
\begin{minipage}[t]{\wd\sideBtab}\offinterlineskip
\usebox{\sideBcap}\par
\ifdim\sideH>\sideHB\vskip\dimexpr\sideH-\sideHB\relax\fi
\usebox{\sideBtab}
\end{minipage}
\end{table*}

\section{Experiments}
\label{sec:experiments}

We describe the experimental setup in Sec.~\ref{sec:setup}, compare \method\ with state-of-the-art methods in Sec.~\ref{sec:sota}, and present ablation studies and analysis in Sec.~\ref{sec:ablation}.

\subsection{Experimental Setup}
\label{sec:setup}

\paragraph{Datasets.}
\method\ is trained on 1.54M images, consisting of 1.46M images from AnyWord-3M~\cite{anytext} and 75.6K real-scene images from the training splits of CTW~\cite{ctw}, TextOCR~\cite{textocr}, HierText~\cite{hiertext}, ReCTS~\cite{rects}, LSVT~\cite{lsvt}, MLT2019~\cite{mlt2019}, Total-Text~\cite{totaltext}, and STIPLAR~\cite{stellar}, after annotation filtering, duplicate removal, and resolution normalization.
For ablations and zero-shot multilingual evaluation, we additionally train \method-Bi on a balanced 100K Chinese--English subset that contains no other script.
Dataset construction details are provided in the supplement.

\paragraph{Benchmarks.}
We evaluate in three settings.
(i)~Scene text generation on AnyText-benchmark~\cite{anytext}, with 1,000 English and 1,000 Chinese images containing 4,181 and 2,092 text lines.
(ii)~Scene text editing on the TextOCR~\cite{textocr}, Total-Text~\cite{totaltext}, and ReCTS~\cite{rects} test sets released with TextFlux~\cite{textflux}, with 3,112/265/1,622 images and 19,531/1,241/4,858 text regions.
On Total-Text we further report three overlapping subsets that strain glyph-image conditioning, namely curved text, small text whose short side is below 16 pixels, and cluttered images with six or more text regions.
(iii)~Zero-shot scene text editing in unseen scripts on STIPLAR~\cite{stellar}, with 385 Japanese, 2,079 Korean, and 981 Arabic examples.
\method\ is evaluated in (i) and (ii), and \method-Bi in (iii).

\paragraph{Metrics.}
Following AnyText~\cite{anytext}, text accuracy is measured by sentence accuracy (ACC) and normalized edit distance (NED) of an OCR model on the generated text regions, and visual quality by FID~\cite{fid} and LPIPS~\cite{lpips} against the reference images.
On STIPLAR, metrics are computed on the edited text crops with language-specific recognizers, and NED and LPIPS are reported in the supplement.

\paragraph{Baselines.}
We compare \method\ with AnyText~\cite{anytext}, AnyText2~\cite{anytext2}, DreamText~\cite{dreamtext}, TextSSR~\cite{textssr}, FLUX-Text~\cite{fluxtext}, TextFlux~\cite{textflux}, and STELLAR~\cite{stellar} using their official checkpoints and sampling settings, and take the results of earlier methods in Tab.~\ref{tab:anytext} from their papers.
The closest region-independent conditioning methods cannot be run under our editing protocol, as EasyText~\cite{easytext} and GlyphPrinter~\cite{glyphprinter} release only prompt-and-layout generation without a source scene or mask, and GlyphAnchor~\cite{glyphanchor} has no public code.

\paragraph{Implementation details.}
\method\ is built on FLUX.1-Fill-dev~\cite{flux2024} and trained as described in Sec.~\ref{sec:train} with AdamW~\cite{adamw}, weight decay 0.01, and gradient clipping 0.5.
It is trained for 5 epochs (about 60K updates) with a global batch size of 128 on 32 H200 GPUs, where the learning rate decays from $5\times10^{-5}$ to $1\times10^{-5}$ with a cosine schedule in the first epoch and stays constant afterwards.
Atlas cells are rendered at 64 pixels in the first epoch and with random resolution augmentation in the remaining four.
\method-Bi and all ablation models are trained for 12.5K iterations with a batch size of 8 on eight H200 GPUs with a constant learning rate of $2\times10^{-5}$ after warm-up, with atlas cells fixed at 64 pixels unless stated otherwise.
All results are sampled with 10 Euler steps and a guidance scale of 1.0.

\begin{table*}[t]
\centering\small
\caption{Ablation on glyph references and attention.
All variants follow the training recipe of \method-Bi, and the last row is \method-Bi.
Full attention lets every token attend to every atlas cell.
}

\label{tab:condition}
\vspace{-4pt}
\setlength{\tabcolsep}{2.5pt}
\begin{tabular*}{\textwidth}{@{\extracolsep{\fill}}ccc*{9}{r}@{}}
\toprule
\multicolumn{2}{@{}c}{Glyph Reference} & \multirow{2}{*}[-0.60ex]{Attention} & \multicolumn{2}{c}{AnyText-benchmark} & \multicolumn{4}{c}{Total-Text} & \multicolumn{3}{c}{STIPLAR} \\
\cmidrule(r){1-2}\cmidrule(lr){4-5}\cmidrule(lr){6-9}\cmidrule(l){10-12}
Image & Atlas & & English & Chinese & All & Curved & Small & Cluttered & Japanese & Korean & Arabic \\
\midrule
$\checkmark$ & -- & -- & 83.90 & 67.93 & 51.33 & 50.86 & 17.65 & 48.51 & 60.26 & 37.90 & 19.27 \\
-- & $\checkmark$ & Anchored & 80.30 & 74.28 & 50.53 & 48.41 & 22.43 & 47.79 & 67.01 & 58.25 & \textbf{39.96} \\
\addlinespace[2pt]
$\checkmark$ & $\checkmark$ & Full & 84.64 & 75.76 & 54.25 & 52.08 & 21.69 & 52.12 & \textbf{70.91} & 65.22 & 32.93 \\
$\checkmark$ & $\checkmark$ & Anchored & \textbf{85.89} & \textbf{76.00} & \textbf{56.97} & \textbf{53.91} & \textbf{27.21} & \textbf{56.08} & 67.53 & \textbf{66.96} & 39.55 \\
\bottomrule
\end{tabular*}
\end{table*}
\begin{figure*}[t]\centering
\includegraphics[width=\textwidth]{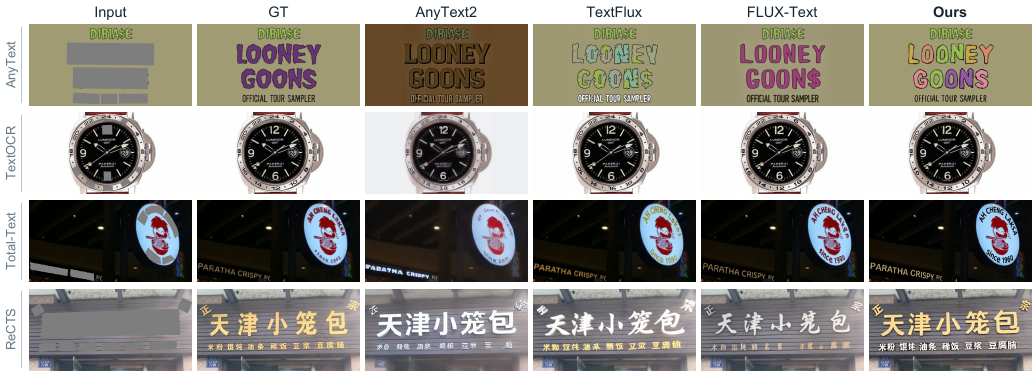}
\vspace{-7pt}
\caption{
Qualitative comparison on evaluated benchmarks.
Gray marks the regions to be filled.
More comparisons are in the supplementary.
``GT'' is the abbreviation for Ground Truth, and all columns of a row show the same crop.
}
\label{fig:qualitative}
\end{figure*}

\subsection{Comparison with State-of-the-Art Methods}
\label{sec:sota}

\paragraph{Scene text generation.}
Tab.~\ref{tab:anytext} presents the results on AnyText-benchmark.
\method\ achieves the best ACC, NED, and FID in both languages, surpassing FLUX-Text, the strongest baseline, by 4.5 ACC points in English and 6.6 in Chinese, with English LPIPS slightly behind FLUX-Text.
The gain is larger in Chinese, whose characters carry the most complex stroke structure, which agrees with our analysis in Sec.~\ref{sec:intro} as the propose atlas contributes the most where the glyph image provides the weakest stroke reference.

\paragraph{Scene text editing.}
Tab.~\ref{tab:scene} presents the results on three real-scene benchmarks, where text regions are small, irregular, and numerous, with 6.3 regions per image on TextOCR.
\method\ achieves the best ACC and NED on all three benchmarks, exceeding the best baseline by 5.4, 9.1, and 19.0 ACC points on TextOCR, Total-Text, and ReCTS, with FID comparable to FLUX-Text and TextFlux.
The gain peaks on ReCTS, which pairs Chinese characters with the small regions of real scenes, the two conditions under which the glyph image is the most under-resolved.

\paragraph{Zero-shot editing in unseen scripts.}
Tab.~\ref{tab:stiplar} presents the results on STIPLAR with \method-Bi, whose training data contain no Korean or Arabic.
Among methods without language-specific training, \method-Bi is the most accurate in all three scripts, exceeding the best such baseline by 19.7, 33.1, and 33.1 ACC points.
It also surpasses STELLAR in Japanese, while STELLAR stays ahead in Korean and Arabic with a separate model fine-tuned on each language.
Notably, FLUX-Text and TextFlux share the backbone with \method-Bi and are trained on far more data, so the gap stems from the glyph reference rather than the backbone or the data, which we verify in Sec.~\ref{sec:ablation}.

\paragraph{Qualitative results.}
Fig.~\ref{fig:qualitative} compares \method\ with AnyText2, TextFlux, and FLUX-Text on the four benchmarks.
The baselines misspell or garble text in every row, such as ``GOON\$'' on the album cover, where the dollar sign is copied from the ``DIBIA\$E'' header, ``16'' and ``18'' for the 10 and 12 on the watch bezel, ``1990'' and ``1900'' on the round sign, and wrong or missing characters in the small menu line of the shop sign, whereas \method\ renders these targets correctly and keeps the hand-drawn outlines, the engraved bezel, and the raised letters of their carriers.

\begin{table}[t]
\centering\footnotesize
\caption{Atlas budget on AnyText-benchmark.
Each row is one inference setting: cells fixed at 64 or 48 pixels, or set by the token allocator (Alloc.).
Fixed and Random are models trained without and with random resolution augmentation.}
\label{tab:budget}
\vspace{-4pt}
\setlength{\tabcolsep}{2.5pt}
\begin{tabular*}{\linewidth}{@{\extracolsep{\fill}}lrrrrrrrr@{}}
\toprule
\multirow{3}{*}[-0.6ex]{Inference} & \multicolumn{4}{c}{English} & \multicolumn{4}{c}{Chinese} \\
\cmidrule(lr){2-5}\cmidrule(l){6-9}
 & \multicolumn{2}{c}{Atlas length} & \multicolumn{2}{c}{ACC $\uparrow$} & \multicolumn{2}{c}{Atlas length} & \multicolumn{2}{c}{ACC $\uparrow$} \\
\cmidrule(lr){2-3}\cmidrule(lr){4-5}\cmidrule(lr){6-7}\cmidrule(l){8-9}
 & Mean & Max & Fixed & Random & Mean & Max & Fixed & Random \\
\midrule
64 px  & 291 & 784 & 88.48 & 88.68 & 175 & 1,200 & 77.77 & 77.92 \\
48 px  & 164 & 441 & 87.91 & 88.10 &  99 &   675 & 72.56 & 74.81 \\
Alloc. & 267 & 512 & 88.15 & 88.22 & 194 &   794 & 75.48 & 76.86 \\
\bottomrule
\end{tabular*}
\end{table}

\subsection{Ablation Studies and Analysis}
\label{sec:ablation}

All models in this section follow the recipe of \method-Bi in Sec.~\ref{sec:setup} and differ from it in one factor at a time.
We ablate the glyph reference, locate the source of zero-shot transfer, and examine the atlas budget.

\paragraph{Glyph references.}
The first two rows of Tab.~\ref{tab:condition} use a single reference and verify that the two are complementary.
The glyph image leads on English and curved text by 3.6 and 2.5 ACC points, where layout is the main difficulty and Latin strokes survive a scene-resolution rendering.
While the atlas leads on Chinese and small text by 6.4 and 4.8 points and on all three STIPLAR scripts, where the glyph image is under-resolved.
This split matches the two roles identified in Sec.~\ref{sec:intro}, and combining the two in the last row inherits the strength of both, improving on the glyph image alone by 2.0 points in English, 8.1 in Chinese, 9.6 on small text, and 7.6 on cluttered images.

\paragraph{Atlas-anchored attention.}
The last two rows hold both references fixed and replace atlas-anchored attention with full attention, which lets every token attend to every cell.
Anchoring improves ACC on every Total-Text subset, most on small text (21.69 vs.\ 27.21) and cluttered images (52.12 vs.\ 56.08), and by 6.6 points in Arabic.
These are the cases where glyphs are tiny and cells of several regions coexist, so a region under full attention can draw strokes from the cells of its neighbors, which is the interference the mask in Eq.~\eqref{eq:binding} removes.
The same effect explains the large margin of \method\ on TextOCR in Tab.~\ref{tab:scene}, whose images carry the most regions.
The one exception is Japanese, where full attention leads by 3.4 points.

\paragraph{Source of zero-shot transfer.}
The four models share the recipe, so what they know about Korean and Arabic comes either from the pretrained prior or from the glyph reference at test time.
The glyph image alone reaches 37.90 and 19.27 ACC, the most the prior recovers from a scene-resolution rendering, whereas the atlas alone reaches 58.25 and 39.96, and the combination 66.96 and 39.55.
As anticipated, the transfer comes from the atlas, whose cells present the strokes of an unseen glyph at a resolution the model can copy, which the VAE encodes without a script-specific vocabulary and atlas-anchored attention delivers to the right region.

\paragraph{Atlas budget.}
Random resolution augmentation lets the atlas budget be cut at inference with a smaller accuracy loss, mainly in Chinese.
A character cell costs 16 tokens at the default 64 pixels and 9 at 48 pixels, so 48-pixel cells shorten every atlas by 44\%.
The token allocator of Sec.~\ref{sec:train} instead bounds text-heavy scenes, shortening the longest atlas from 784 to 512 tokens in English and from 1,200 to 794 in Chinese, and changes the mean by under 11\% because it also raises the cells of short atlases.
Tab.~\ref{tab:budget} applies both cuts to two models continued from the same checkpoint, one with random resolution augmentation and one with cells fixed at 64 pixels.
The two are within 0.2 ACC points at 64 pixels, and English stays within 0.6 points for both under either cut.
In Chinese, augmentation reduces the loss from 5.2 to 3.1 points with 48-pixel cells and from 2.3 to 1.1 with the allocator.
The full train--test grid of cell heights and the complete budget results are in the supplementary.

\section{Conclusion}
\label{sec:conclusion}

We propose \method, a scene text synthesis framework, to address the challenge of rendering stroke-accurate text when the glyph image is under-resolved for small, dense, or structurally complex text.
We first introduce the glyph atlas, which renders each character or word in a separate cell at a resolution independent of its target region and supplies fine-grained stroke references.
Building on this, we design atlas-anchored attention to bind each atlas cell to its own text region, so that the glyph image anchors the layout while the atlas refines the strokes.
In addition, random resolution augmentation of atlas cells makes the atlas token budget adjustable at inference.
Extensive experiments demonstrate that \method\ outperforms existing methods in text accuracy on four scene text benchmarks and transfers zero-shot to scripts unseen in training, demonstrating significant potential in scene text synthesis for low-resource languages.

\paragraph{Limitation.}
Although \method\ achieves high text accuracy, its atlas cells are rendered in a uniform font and thus specify strokes rather than typography, leaving occasional errors in font weight and color.
Also,
the appended atlas tokens inevitably incurs extra inference cost over glyph-image-only conditioning despite the introduced adjustable budget.
We regard these as promising directions for future works.

{\small\bibliographystyle{ieeenat_fullname}\bibliography{references}}

\clearpage
\appendix
\let\paragraph\originalparagraph
\setcounter{table}{0}
\setcounter{figure}{0}
\renewcommand{\thetable}{S\arabic{table}}
\renewcommand{\thefigure}{S\arabic{figure}}
\renewcommand{\theHtable}{supp.\arabic{table}}
\renewcommand{\theHfigure}{supp.\arabic{figure}}
\twocolumn[{\centering\Large\bfseries Supplementary Material\par\vspace{1em}}]
This supplement is organized as follows.
Sec.~\ref{sec:supp_setup} details the experimental setup: the construction of the training data, the evaluation protocol, and the implementation of \method\ and of the compared methods.
Sec.~\ref{sec:supp_related} extends the related work of the main paper.
Secs.~\ref{sec:supp_full}--\ref{sec:supp_budget} report the full results behind the abbreviated tables of the main paper: the comparison with state-of-the-art methods (Sec.~\ref{sec:supp_full}), the ablation studies together with two further design ablations (Sec.~\ref{sec:supp_ablation}), and the atlas budget study with its inference cost (Sec.~\ref{sec:supp_budget}).
Sec.~\ref{sec:supp_qualitative} closes with more qualitative results on scene text generation, scene text editing, and out-of-distribution case studies.
Unless stated otherwise, every ablation model follows the recipe of \method-Bi in Sec.~4.1 of the main paper and differs from it in one factor at a time, and ACC denotes sentence accuracy on AnyText-benchmark.

\section{Experimental Setup}
\label{sec:supp_setup}

We describe how the training data are built (Sec.~\ref{sec:supp_data}), how every reported number is computed (Sec.~\ref{sec:supp_protocol}), and how \method\ and the compared methods are implemented and run (Sec.~\ref{sec:supp_impl}).

\subsection{Dataset Construction}
\label{sec:supp_data}

\method\ is trained on a pool of 1.54M real-scene images, and \method-Bi and all ablations on a balanced 100K Chinese--English subset.
We describe how each is built.

\paragraph{Main training pool.}
The pool combines 1.46M images from the deduplicated AnyWord-3M~\cite{anytext} with 75.6K additional real-scene images drawn from the training splits of CTW~\cite{ctw}, TextOCR~\cite{textocr}, HierText~\cite{hiertext}, ReCTS~\cite{rects}, LSVT~\cite{lsvt}, MLT2019~\cite{mlt2019}, Total-Text~\cite{totaltext}, and STIPLAR~\cite{stellar}.
For the additional images, we take at most two $512\times512$ crops per source image, at the original scale and, optionally, with up to $2\times$ enlargement, and never resize a crop after cropping.
A region is kept only if its smallest character is at least 20 pixels, its local contrast and sharpness pass fixed thresholds, and it overlaps an ignore region by at most 5\%.
Pixel-level duplicates are matched against AnyWord-3M and resolved in its favor.
Original transcriptions are retained throughout.

\paragraph{Bilingual subset.}
The subset used for \method-Bi and the ablations contains 50K Chinese-annotated and 50K Latin-only images, selected deterministically from the pool so that the choice is reproducible and contains no script other than Chinese and Latin.
It thus isolates the effect of each design factor from the long multilingual tail of the full pool, and leaves Japanese, Korean, and Arabic entirely unseen for the zero-shot evaluation of Sec.~4.2 of the main paper.

\subsection{Evaluation Protocol}
\label{sec:supp_protocol}

\paragraph{Text accuracy.}
On AnyText-benchmark we score the generated regions with the ModelScope DuGuang recognizer~\cite{ppocrv3}, following the AnyText protocol~\cite{anytext}; the PP-OCRv3 recognizer that serves as the training perceptual teacher is used only as a diagnostic and never produces a reported accuracy.
A region is cropped from the output with its annotated polygon, rotated upright when its height exceeds $1.2\times$ its width, resized to a height of 48 pixels with width capped at 320, and padded to width 320.
Sentence accuracy (ACC) is the fraction of regions whose recognized string equals the target exactly, averaged over all scored regions rather than over images, with no case folding or punctuation stripping.
Normalized edit distance is $\mathrm{NED}=1-\mathrm{Lev}(\hat y,y)/\max(|\hat y|,|y|)$, higher being better, averaged over the same regions.
On STIPLAR we use the language-specific PP-OCRv4 recognizers for Japanese, Korean, and Arabic, each run on the edited crop; for Arabic we keep the logical character order and additionally remove spaces, matching the STELLAR evaluation~\cite{stellar}.
ACC is a recognizer-dependent estimate of legibility, not a per-stroke guarantee.

\paragraph{Image quality.}
FID~\cite{fid} uses the 2{,}048-dimensional Inception features and LPIPS~\cite{lpips} the AlexNet features.
On AnyText-benchmark and the real-scene editing benchmarks both are computed on the full images; on STIPLAR, where only a crop is edited, both are computed on the edited crops, with LPIPS inputs resized to $256\times256$.

\paragraph{Total-Text subsets.}
The three Total-Text subsets reported in the main paper are defined on the TextFlux-released view of 265 images with 1{,}241 edited regions.
Curved collects the regions annotated as curved; small collects the regions whose short side is below 16 pixels on the $512\times512$ canvas; cluttered collects the regions of images with six or more edited regions.
The subsets overlap, and a region is scored under every subset it satisfies.

\subsection{Implementation Details}
\label{sec:supp_impl}

We complement the implementation details of Sec.~4.1 of the main paper with the components that are not specified there.

\paragraph{Glyph image and prompt.}
Following FLUX-Text~\cite{fluxtext}, each target string is drawn into the glyph image in Arial Unicode, centered in the minimum-area rectangle of its region and rotated with it.
Its font size is 0.8 of the rectangle's short side, reduced when the string would overflow the long side, and spaces are inserted between characters to spread a short string along the rectangle, after which the glyph image is binarized.
The atlas cells are rendered in the same font.
During training, the prompt is the image caption followed by one of ten template phrases and the target strings, \eg, ``a bookstore, that reads BOOKS .'', and at most eight text regions are used per image.

\paragraph{Glyph atlas.}
Text units are formed by a deterministic rule: each Han character is a unit, each word of a letter-based script such as Latin or Arabic is a word unit, with digits, internal apostrophes, and hyphens kept inside it, combining marks stay with their base character, and spaces produce no cell.
Japanese kana and Korean syllables are character units.
A character is rendered white on black into a $56\times56$ box centered in a $64\times64$ cell.
A word is rendered at an ink height of at most 56 pixels into a strip whose width is the smallest multiple of 64 pixels exceeding the ink width by at least 8 pixels, and the strip is split into $64\times64$ tiles that all belong to the same unit.
Each tile is encoded with the posterior mean of the frozen VAE, normalized with the shift and scale of the VAE, and packed into $2\times2$ latent patches, which gives 16 tokens of dimension 64.
A cell of height $h$ and width $w$ thus costs $hw/256$ tokens, \ie, 9, 16, and 25 tokens for a square cell at 48, 64, and 80 pixels, and cells at other heights are re-rendered from the font outlines rather than resized from a 64-pixel bitmap.

\paragraph{Atlas tokens and attention.}
Atlas tokens are projected to the model width of 3{,}072 by a bias-free linear layer initialized as described in Sec.~3.2 of the main paper.
A two-layer MLP, whose output layer is initialized to zero, adds an embedding of the position of each token inside its cell together with the normalized unit index $j$ and the number of units in its region.
These two modules are the only new parameters, 0.59M in total against 11.9B in the backbone.
Atlas tokens are appended to the glyph-image tokens and share their modulation, and their rotary positions place each region in its own band of rows on a canvas separate from the scene grid, with the cells of a region in reading order.
Atlas-anchored attention is computed without a dense mask.
Queries are grouped by the set of regions that their patch overlaps, where a $16\times16$-pixel patch overlaps a region if any of its pixels lies inside the region, and each group runs standard scaled dot-product attention over the keys that Eq.~(5) of the main paper allows it.

\paragraph{Training.}
We use AdamW~\cite{adamw} with $\beta_1=0.9$, $\beta_2=0.999$, and $\epsilon=10^{-8}$, keep the master weights and optimizer states in FP32 with BF16 computation, and shard the model with FSDP.
Every GPU processes one image per step; for \method, gradients are accumulated over four steps on the 32 GPUs to reach the global batch size of 128.
The prompt and the glyph image are each dropped with probability 0.1, and the atlas is dropped whenever the glyph image is.
For $\mathcal L_{\rm OCR}$, each region is cropped from $\tilde I$ and $I$ along the minimum bounding rectangle of its mask, resized and padded to the $48\times320$ input of the frozen PP-OCRv3 recognizer~\cite{ppocrv3}, and compared through the features that the recognizer feeds to its CTC classifier.
The reference branch runs without gradients, and no crop is filtered by its transcription or by the recognizer's confidence.

\paragraph{Inference.}
All benchmark images are generated at $512\times512$ with 10 Euler steps, a guidance scale of 1.0, seed 42, and BF16, with 64-pixel cells unless stated otherwise.

\paragraph{Compared methods.}
For the results we measure, each compared method runs from its official checkpoint with its recommended sampling: AnyText~\cite{anytext} with 20 steps and guidance 9, AnyText2~\cite{anytext2} with 20 steps and guidance 7.5, FLUX-Text~\cite{fluxtext} with 28 steps and guidance 3.5, and TextFlux~\cite{textflux} with 50 steps and guidance 30.
TextSSR~\cite{textssr} edits the regions of an image one after another with its native 20-step local editor, and on STIPLAR it uses its released font for Japanese and Arial Unicode MS for Korean and Arabic.
DreamText~\cite{dreamtext} accepts only ASCII targets, so its results in Tabs.~1 and~2 of the main paper are computed, with 50 steps and guidance 5, on the images it supports: 950 English and 91 Chinese images of AnyText-benchmark and 2{,}539, 254, and 75 images of TextOCR, Total-Text, and ReCTS.
STELLAR~\cite{stellar} uses its released stage-2 model for each language with 50 inversion and 50 editing steps at guidance 2, on its native $256\times256$ text crops.
On STIPLAR, \method, AnyText2, FLUX-Text, and TextFlux receive a square scene crop centered on the source text box, whose side is four times the longer side of the box, clamped to 256--1{,}024 pixels, and resized to $512\times512$; TextFlux stacks its glyph canvas above this crop into a $512\times1024$ input.

\section{Extended Related Work}
\label{sec:supp_related}

\subsection{Text-to-Image Diffusion Models}
Diffusion models~\cite{ddpm,ldm} have become the dominant paradigm for visual generation.
Architecturally, they have evolved from UNet-based latent diffusion~\cite{ldm,sdxl} to diffusion transformers (DiTs)~\cite{dit} trained with flow matching~\cite{flowmatching,rectifiedflow}, as represented by SD3~\cite{sd3} and FLUX~\cite{flux2024}, which treat the image as a token sequence and condition on the prompt through joint attention.
Conditioned on a text prompt alone, these models offer weak control over spatial layout and over exact symbol identity, and visual text exposes both weaknesses at once, since a word must appear at a specific place with a specific spelling~\cite{characteraware}.
The common remedy is an additional image condition that states what the prompt cannot.
In UNet-based models it is injected through an auxiliary branch~\cite{controlnet}; in DiTs it is encoded by the same VAE as the image and concatenated with the noisy latent into a unified token sequence, so that the two interact through ordinary self-attention~\cite{ominicontrol,icedit,fluxkontext}.
The latter design is relevant to us for two reasons: the condition needs no dedicated encoder, and an attention mask can restrict which condition tokens each image token sees.
Scene text synthesis follows this recipe with a pre-rendered glyph image as the condition, and \cref{subsec:supp_scene} discusses how this condition is built and where it falls short.

\begin{table}[t]\centering\footnotesize\setlength{\tabcolsep}{4pt}
\caption{\textbf{Zero-shot multilingual scene text editing on STIPLAR}, completing Tab.~3 of the main paper with NED and LPIPS.
$\ddagger$: a separate model tuned per language.
Best among methods without language-specific tuning is in \textbf{bold}.}
\label{tab:stiplar_full}
\begin{tabular}{llrrr}\toprule
Script & Method & ACC$\uparrow$ & NED$\uparrow$ & LPIPS$\downarrow$ \\
\midrule
\multirow{6}{*}{Japanese}
 & AnyText2~\cite{anytext2}  & 47.79 & 0.6599 & 0.4745 \\
 & TextSSR~\cite{textssr}    & 40.26 & 0.5790 & 0.4786 \\
 & FLUX-Text~\cite{fluxtext} & 46.49 & 0.6979 & 0.3806 \\
 & TextFlux~\cite{textflux}  & 46.23 & 0.6811 & \textbf{0.3502} \\
 & STELLAR$^\ddagger$~\cite{stellar} & 43.38 & 0.6354 & 0.3878 \\
 & \method-Bi & \textbf{67.53} & \textbf{0.8302} & 0.3853 \\
\midrule
\multirow{6}{*}{Korean}
 & AnyText2~\cite{anytext2}  & 29.82 & 0.5886 & 0.4519 \\
 & TextSSR~\cite{textssr}    & 19.14 & 0.3944 & 0.4954 \\
 & FLUX-Text~\cite{fluxtext} & 20.83 & 0.5275 & 0.3918 \\
 & TextFlux~\cite{textflux}  & 33.86 & 0.6720 & 0.3423 \\
 & STELLAR$^\ddagger$~\cite{stellar} & 80.47 & 0.9117 & 0.2526 \\
 & \method-Bi & \textbf{66.96} & \textbf{0.8560} & \textbf{0.3434} \\
\midrule
\multirow{6}{*}{Arabic}
 & AnyText2~\cite{anytext2}  &  0.10 & 0.1645 & 0.5444 \\
 & TextSSR~\cite{textssr}    &  0.20 & 0.2063 & 0.5658 \\
 & FLUX-Text~\cite{fluxtext} &  2.65 & 0.3667 & 0.4892 \\
 & TextFlux~\cite{textflux}  &  6.42 & 0.5205 & 0.4484 \\
 & STELLAR$^\ddagger$~\cite{stellar} & 55.86 & 0.8944 & 0.3858 \\
 & \method-Bi & \textbf{39.55} & \textbf{0.8050} & \textbf{0.4383} \\
\bottomrule
\end{tabular}
\end{table}

\begin{table*}[t]\centering\footnotesize\setlength{\tabcolsep}{2pt}
\caption{\textbf{Full metrics of the glyph-reference and attention ablation}, completing Tab.~4 of the main paper.
All four models follow the training recipe of \method-Bi, and the last row of each block is \method-Bi.
Full attention lets every token attend to every atlas cell.
Total-Text uses 265 images with 1,241 edited regions, and its FID is computed on full images of the whole set only.
STIPLAR FID and LPIPS are computed on the edited text crops.
Best results are in \textbf{bold}.}
\label{tab:condition_full}
\textit{AnyText-benchmark}\par\smallskip
\begin{tabular*}{\textwidth}{@{\extracolsep{\fill}}ccc*{8}{r}@{}}
\toprule
\multicolumn{2}{@{}c}{Glyph Reference} & \multirow{2}{*}[-0.60ex]{Attention} & \multicolumn{4}{c}{English} & \multicolumn{4}{c}{Chinese} \\
\cmidrule(r){1-2}\cmidrule(lr){4-7}\cmidrule(lr){8-11}
Image & Atlas &  & ACC$\uparrow$ & NED$\uparrow$ & FID$\downarrow$ & LPIPS$\downarrow$ & ACC$\uparrow$ & NED$\uparrow$ & FID$\downarrow$ & LPIPS$\downarrow$ \\
\midrule
$\checkmark$ & -- & -- & 83.90 & 0.9326 & 12.77 & 0.0858 & 67.93 & 0.8414 & 13.37 & 0.0706 \\
-- & $\checkmark$ & Anchored & 80.30 & 0.9078 & 12.99 & 0.0877 & 74.28 & 0.8605 & \textbf{13.30} & 0.0696 \\
$\checkmark$ & $\checkmark$ & Full & 84.64 & 0.9378 & 12.75 & 0.0842 & 75.76 & 0.8713 & 13.36 & \textbf{0.0669} \\
$\checkmark$ & $\checkmark$ & Anchored & \textbf{85.89} & \textbf{0.9430} & \textbf{12.38} & \textbf{0.0829} & \textbf{76.00} & \textbf{0.8734} & 13.35 & 0.0678 \\
\bottomrule
\end{tabular*}
\par\medskip\textit{Total-Text}\par\smallskip
\begin{tabular*}{\textwidth}{@{\extracolsep{\fill}}ccc*{13}{r}@{}}
\toprule
\multicolumn{2}{@{}c}{Glyph Reference} & \multirow{2}{*}[-0.60ex]{Attention} & \multicolumn{4}{c}{All} & \multicolumn{3}{c}{Curved} & \multicolumn{3}{c}{Small} & \multicolumn{3}{c}{Cluttered} \\
\cmidrule(r){1-2}\cmidrule(lr){4-7}\cmidrule(lr){8-10}\cmidrule(lr){11-13}\cmidrule(lr){14-16}
Image & Atlas &  & ACC$\uparrow$ & NED$\uparrow$ & FID$\downarrow$ & LPIPS$\downarrow$ & ACC$\uparrow$ & NED$\uparrow$ & LPIPS$\downarrow$ & ACC$\uparrow$ & NED$\uparrow$ & LPIPS$\downarrow$ & ACC$\uparrow$ & NED$\uparrow$ & LPIPS$\downarrow$ \\
\midrule
$\checkmark$ & -- & -- & 51.33 & 0.6943 & 26.67 & 0.0675 & 50.86 & 0.7111 & 0.0678 & 17.65 & 0.3532 & 0.0677 & 48.51 & 0.6680 & 0.0724 \\
-- & $\checkmark$ & Anchored & 50.53 & 0.6957 & 27.25 & 0.0678 & 48.41 & 0.6992 & 0.0680 & 22.43 & 0.4175 & 0.0676 & 47.79 & 0.6745 & 0.0728 \\
$\checkmark$ & $\checkmark$ & Full & 54.25 & 0.7028 & 26.68 & 0.0656 & 52.08 & 0.7033 & 0.0659 & 21.69 & 0.3928 & 0.0658 & 52.12 & 0.6836 & \textbf{0.0700} \\
$\checkmark$ & $\checkmark$ & Anchored & \textbf{56.97} & \textbf{0.7389} & \textbf{26.51} & \textbf{0.0652} & \textbf{53.91} & \textbf{0.7362} & \textbf{0.0654} & \textbf{27.21} & \textbf{0.4718} & \textbf{0.0654} & \textbf{56.08} & \textbf{0.7286} & 0.0702 \\
\bottomrule
\end{tabular*}
\par\medskip\textit{STIPLAR}\par\smallskip
\begin{tabular*}{\textwidth}{@{\extracolsep{\fill}}ccc*{12}{r}@{}}
\toprule
\multicolumn{2}{@{}c}{Glyph Reference} & \multirow{2}{*}[-0.60ex]{Attention} & \multicolumn{4}{c}{Japanese} & \multicolumn{4}{c}{Korean} & \multicolumn{4}{c}{Arabic} \\
\cmidrule(r){1-2}\cmidrule(lr){4-7}\cmidrule(lr){8-11}\cmidrule(lr){12-15}
Image & Atlas &  & ACC$\uparrow$ & NED$\uparrow$ & FID$\downarrow$ & LPIPS$\downarrow$ & ACC$\uparrow$ & NED$\uparrow$ & FID$\downarrow$ & LPIPS$\downarrow$ & ACC$\uparrow$ & NED$\uparrow$ & FID$\downarrow$ & LPIPS$\downarrow$ \\
\midrule
$\checkmark$ & -- & -- & 60.26 & 0.7633 & 84.96 & 0.3924 & 37.90 & 0.6764 & 37.75 & 0.3718 & 19.27 & 0.6337 & 66.68 & 0.4567 \\
-- & $\checkmark$ & Anchored & 67.01 & 0.8311 & \textbf{82.74} & 0.3939 & 58.25 & 0.8002 & 39.19 & 0.3648 & \textbf{39.96} & 0.8042 & 67.28 & \textbf{0.4348} \\
$\checkmark$ & $\checkmark$ & Full & \textbf{70.91} & \textbf{0.8522} & 84.84 & 0.3881 & 65.22 & 0.8504 & \textbf{36.76} & 0.3478 & 32.93 & 0.7582 & \textbf{63.59} & 0.4449 \\
$\checkmark$ & $\checkmark$ & Anchored & 67.53 & 0.8302 & 84.09 & \textbf{0.3853} & \textbf{66.96} & \textbf{0.8560} & 38.99 & \textbf{0.3434} & 39.55 & \textbf{0.8050} & 68.10 & 0.4383 \\
\bottomrule
\end{tabular*}
\end{table*}

\subsection{Scene Text Synthesis}
\label{subsec:supp_scene}
Scene text synthesis covers scene text generation and editing, both rendering legible text in specified regions of a scene image while keeping it coherent with the surroundings~\cite{anytext,fluxtext}.
Early methods composite rendered text onto natural images for data synthesis~\cite{synthtext} or edit existing text with GANs by transferring the style of the source text to the target string~\cite{srnet}.
Diffusion models have since taken over.
UNet-based methods formulate the task as inpainting and add auxiliary glyph conditions, such as character-level segmentation masks~\cite{textdiffuser}, glyph images fed through a control branch~\cite{glyphcontrol,anytext}, or glyph features from a text or OCR encoder~\cite{diffste,diffute}.
Recent methods fine-tune DiTs instead, either with the glyph image as an extra input channel~\cite{fluxtext} or by concatenating it spatially with the scene so that the backbone copies glyphs through in-context attention~\cite{textflux}.
Across these models, the prevailing condition is a glyph image, where the target text is pre-rendered in a canonical font at its target locations on a scene-sized canvas~\cite{glyphcontrol,anytext,fluxtext,textflux}.
Its appeal is that one image, pixel-aligned with the scene, serves as both spatial anchor and stroke reference: the model knows where each glyph belongs and roughly what it looks like.
Its cost is that a scene-aligned reference inherits the scene's per-region resolution, so each glyph receives as many pixels as its region allows rather than as its strokes require.
Small, dense, and structurally complex text is accordingly the commonly reported failure case, with the glyph image itself identified as the cause~\cite{postermaker,textflux}.
This is the dilemma of the area: a scene-aligned reference anchors well but describes strokes poorly, while a region-independent reference describes strokes well but must be bound back to the scene.
Existing work restores the binding along two routes.

The first route injects glyph embeddings from OCR recognizers or glyph-aligned text encoders, either in place of the glyph image or alongside it~\cite{anytext,anytext2,udifftext,glyphbyt5,postermaker}.
Embeddings are compact and independent of region size, but they have three drawbacks that mirror the issues raised in Sec.~1 of the main paper.
They require an additional encoder whose character set bounds the glyphs it can describe, so unseen characters are poorly covered~\cite{anytext2}.
Their features are trained for recognition, which rewards invariance to exactly the stroke-level variation that rendering must reproduce.
And once injected as prompt tokens they lose correspondence with the region they describe, which is why later work adds position encoders~\cite{anytext2,postermaker} or region-wise cross-attention~\cite{glyphbyt5} to restore it.
The second route, closer to ours, renders each text item as a separate glyph patch at a fixed font size and binds the patch to its region through positional encodings remapped into the target grid~\cite{easytext,glyphanchor} or through an attention mask that pairs each patch with its region~\cite{glyphprinter}.
These methods are developed mainly for text-to-image rendering, where the whole image is synthesized from a prompt and a box layout, and they replace the glyph image rather than complement it.
Layout is thus carried only by indices or masks, no coarse structure is placed in the scene to be refined, and text is missed or corrupted once the binding becomes ambiguous, as when regions overlap or anchoring is removed~\cite{easytext,glyphanchor}.
\method\ takes neither route exclusively.
It keeps the glyph image as spatial anchor and coarse structure, and adds a glyph atlas of cells encoded by the backbone's own VAE, bound region-wise by atlas-anchored attention, solely to refine that structure.
The conditioning thus becomes coarse-to-fine rather than a choice between an aligned-but-coarse and a detailed-but-unbound reference.

\section{Full Comparison Results}
\label{sec:supp_full}

Tab.~3 of the main paper reports ACC and FID for zero-shot multilingual scene text editing on STIPLAR.
Tab.~\ref{tab:stiplar_full} completes it with NED and LPIPS, computed as in Sec.~\ref{sec:supp_protocol} with the inference settings of Sec.~\ref{sec:supp_impl}.
These results demonstrate that:
1)~NED tracks ACC, so \method-Bi leads every method without language-specific tuning on all three scripts by the same margins, most widely in Arabic, where the next such method reaches 0.5205 against our 0.8050;
2)~the one model tuned per language, STELLAR, keeps its NED lead on Korean and Arabic but not on Japanese, which \method-Bi handles zero-shot through the kanji it shares with Chinese;
3)~our LPIPS is on par with the editing baselines despite rendering unseen scripts, so the accuracy gain does not come at the cost of visual quality.

\section{Full Ablation Results}
\label{sec:supp_ablation}

We first complete the ablation on glyph references and attention in Tab.~4 of the main paper (Sec.~\ref{sec:supp_condition}).
We then justify two design choices that the main paper states without evidence, the text unit of the atlas in Sec.~3.2 (Sec.~\ref{sec:supp_granularity}) and the timestep-weighted text perceptual loss in Sec.~3.3 (Sec.~\ref{sec:supp_loss}).

\subsection{Glyph References and Attention}
\label{sec:supp_condition}
Tab.~\ref{tab:condition_full} completes Tab.~4 of the main paper with NED, FID, and LPIPS.
These results demonstrate that:
1)~NED agrees with ACC, as \method-Bi has the best NED wherever it has the best ACC, full attention leads both in Japanese, and in Arabic the atlas-only model leads ACC by 0.4 points while \method-Bi leads NED by 0.0008;
2)~the glyph reference changes text accuracy far more than image quality, as English ACC spans 5.6 points across the four models while FID spans 0.61, and Total-Text ACC spans 6.4 points while LPIPS spans 0.0026;
3)~the image quality of \method-Bi is not traded for accuracy, as it has the best FID and LPIPS in English and on the whole Total-Text set, and in Chinese its FID and LPIPS are within 0.05 and 0.0009 of the best.

\begin{table}[t]
\centering\small\setlength{\tabcolsep}{5pt}
\caption{\textbf{Ablation on cell granularity}, as ACC on AnyText-benchmark. Best results are in \textbf{bold}.}
\label{tab:granularity}
\begin{tabular}{lrr}\toprule
Cell unit & EN & ZH \\
\midrule
Each character             & 83.52 & 74.71 \\
Whole line                 & 84.69 & 75.87 \\
Character / word (ours)    & \textbf{85.89} & \textbf{76.00} \\
\bottomrule
\end{tabular}
\end{table}

\begin{table*}[t]
\centering\small\setlength{\tabcolsep}{4pt}
\caption{\textbf{Ablation on the training objective} on AnyText-benchmark.
All rows use $\mathcal L_{\rm flow}$ plus the listed text supervision with weight $w$; a fixed $w$ is constant throughout training, and $w\cdot 2t$ applies the timestep weight of Eq.~(7) of the main paper.
The last row is \method-Bi. Best results are in \textbf{bold}.}
\label{tab:loss}
\begin{tabular}{llrrrrrrrr}\toprule
 & & \multicolumn{4}{c}{English} & \multicolumn{4}{c}{Chinese} \\
\cmidrule(lr){3-6}\cmidrule(lr){7-10}
Text supervision & $w$ & ACC $\uparrow$ & NED $\uparrow$ & FID $\downarrow$ & LPIPS $\downarrow$ & ACC $\uparrow$ & NED $\uparrow$ & FID $\downarrow$ & LPIPS $\downarrow$ \\
\midrule
None                         & --   & 70.08 & 0.8667 & 14.06 & 0.0789 & 62.19 & 0.7981 & 11.45 & 0.0533 \\
\multirow{3}{*}{ODM}         & 1    & 76.85 & 0.9021 & 11.18 & 0.0733 & 65.30 & 0.8155 & \textbf{10.48} & \textbf{0.0518} \\
                             & 5    & 81.62 & 0.9220 & \textbf{10.87} & \textbf{0.0731} & 69.12 & 0.8409 & 10.95 & 0.0531 \\
                             & 25   & 83.30 & 0.9315 & 11.70 & 0.0756 & 70.27 & 0.8462 & 11.93 & 0.0573 \\
\midrule
\multirow{5}{*}{$\mathcal L_{\rm OCR}$, fixed} & 0.05 & 82.29 & 0.9262 & 13.05 & 0.0798 & 73.52 & 0.8636 & 12.06 & 0.0601 \\
                             & 0.25 & 84.79 & 0.9374 & 12.23 & 0.0825 & 75.62 & 0.8721 & 13.12 & 0.0671 \\
                             & 0.50 & 84.86 & 0.9378 & 13.11 & 0.0883 & 76.24 & 0.8732 & 14.05 & 0.0728 \\
                             & 0.75 & 85.34 & 0.9396 & 12.76 & 0.0872 & \textbf{76.58} & 0.8709 & 14.30 & 0.0730 \\
                             & 1.00 & \textbf{86.16} & \textbf{0.9433} & 13.26 & 0.0903 & 75.91 & \textbf{0.8741} & 14.84 & 0.0770 \\
\midrule
\multirow{4}{*}{$\mathcal L_{\rm OCR}$, timestep-weighted} & $0.10\cdot 2t$ & 83.85 & 0.9339 & 12.50 & 0.0817 & 75.72 & 0.8713 & 12.21 & 0.0631 \\
                             & $0.15\cdot 2t$ & 85.15 & 0.9373 & 12.46 & 0.0828 & 75.10 & 0.8721 & 12.69 & 0.0653 \\
                             & $0.20\cdot 2t$ & 85.32 & 0.9397 & 12.81 & 0.0826 & 75.05 & 0.8711 & 12.84 & 0.0664 \\
                             & $0.25\cdot 2t$ & 85.89 & 0.9430 & 12.38 & 0.0829 & 76.00 & 0.8734 & 13.35 & 0.0678 \\
\bottomrule
\end{tabular}
\end{table*}

\subsection{Cell Granularity}
\label{sec:supp_granularity}
Tab.~\ref{tab:granularity} compares three granularities of atlas cells: one cell per character, which splits a Latin word into letters; one cell per text line; and the text unit of Sec.~3.2 of the main paper, a character in character-based scripts and a word in letter-based scripts.
These results demonstrate that:
1)~splitting words into letters is the worst in both languages, 2.4 and 1.3 ACC points behind the text unit, since the letter spacing and joining forms the model must reproduce exist only within a word;
2)~line cells trail the text unit by 1.2 points in English while matching it in Chinese, as a line cell couples all words of a line into one reference and leaves each word-to-stroke correspondence to be resolved inside the cell, whereas word cells make it explicit through the unit index.
The text unit is therefore adopted as a single rule for all scripts.

\begin{table}[t]
\centering\small\setlength{\tabcolsep}{4pt}
\caption{\textbf{Cell height at training (rows) and inference (columns)}, as ACC on AnyText-benchmark.
Heights of 32, 48, 64, 80, and 96 pixels cost 4, 9, 16, 25, and 36 tokens per character cell.
Full metrics for the 48--80 range are in Tab.~\ref{tab:atlas_full}.
Best results are in \textbf{bold}.}
\label{tab:atlas}
\begin{tabular}{lrrrcrrr}\toprule
 & \multicolumn{3}{c}{EN, test px} & & \multicolumn{3}{c}{ZH, test px} \\
\cmidrule(lr){2-4}\cmidrule(lr){6-8}
Train px & 48 & 64 & 80 & & 48 & 64 & 80 \\
\midrule
32 & 84.62 & 84.48 & 83.90 & & 74.33 & 71.94 & 67.21 \\
48 & 85.08 & 84.69 & 84.29 & & \textbf{75.53} & 73.90 & 69.69 \\
64 & \textbf{85.60} & \textbf{85.89} & \textbf{86.25} & & 71.85 & \textbf{76.00} & 75.72 \\
80 & 84.60 & 84.79 & 85.44 & & 60.95 & 73.61 & 75.72 \\
96 & 82.77 & 83.76 & 84.12 & & 48.18 & 70.08 & 75.33 \\
\bottomrule
\end{tabular}
\end{table}

\begin{table*}[t]\centering\small
\caption{\textbf{Full results of the atlas budget study} on AnyText-benchmark, extending Tab.~5 of the main paper with 80-pixel cells, NED, FID, and LPIPS.
Fixed and Random are the models of Tab.~5, trained without and with random resolution augmentation in the fifth epoch.}
\label{tab:supp_budget}
{\setlength{\tabcolsep}{3pt}
\begin{tabular}{llrrrrrrrr}\toprule
 & & \multicolumn{4}{c}{English} & \multicolumn{4}{c}{Chinese} \\
\cmidrule(lr){3-6}\cmidrule(lr){7-10}
Training & Inference & ACC $\uparrow$ & NED $\uparrow$ & FID $\downarrow$ & LPIPS $\downarrow$ & ACC $\uparrow$ & NED $\uparrow$ & FID $\downarrow$ & LPIPS $\downarrow$ \\
\midrule
\multirow{4}{*}{Fixed} & 64 & 88.48 & 0.9552 & 12.00 & 0.0759 & 77.77 & 0.8806 & 11.50 & 0.0574 \\
 & 48 & 87.91 & 0.9536 & 11.93 & 0.0757 & 72.56 & 0.8635 & 11.52 & 0.0593 \\
 & 80 & 88.63 & 0.9557 & 11.96 & 0.0761 & 76.20 & 0.8754 & 11.65 & 0.0585 \\
 & Alloc. & 88.15 & 0.9547 & 12.00 & 0.0758 & 75.48 & 0.8731 & 11.63 & 0.0587 \\
\midrule
\multirow{4}{*}{Random} & 64 & 88.68 & 0.9555 & 12.27 & 0.0756 & 77.92 & 0.8802 & 11.90 & 0.0574 \\
 & 48 & 88.10 & 0.9533 & 12.48 & 0.0756 & 74.81 & 0.8716 & 11.83 & 0.0585 \\
 & 80 & 88.53 & 0.9545 & 12.26 & 0.0758 & 76.96 & 0.8774 & 12.01 & 0.0580 \\
 & Alloc. & 88.22 & 0.9546 & 12.27 & 0.0756 & 76.86 & 0.8771 & 11.95 & 0.0580 \\
\bottomrule\end{tabular}}
\end{table*}

\begin{table}[t]\centering\small
\caption{\textbf{Atlas length per image} on AnyText-benchmark, in tokens, for each inference setting of Tab.~\ref{tab:supp_budget}. The lengths are the same for both models.}
\label{tab:budget_length}
{\setlength{\tabcolsep}{4pt}
\begin{tabular}{lrrcrr}\toprule
 & \multicolumn{2}{c}{English} & & \multicolumn{2}{c}{Chinese} \\
\cmidrule(lr){2-3}\cmidrule(lr){5-6}
Inference & Mean & Max & & Mean & Max \\
\midrule
64 & 291 & 784 & & 175 & 1,200 \\
48 & 164 & 441 & & 99 & 675 \\
80 & 455 & 1,225 & & 274 & 1,875 \\
Alloc. & 267 & 512 & & 194 & 794 \\
\bottomrule\end{tabular}}
\end{table}

\begin{table*}[t]\centering\small
\caption{\textbf{Full metrics of Tab.~\ref{tab:atlas}} for training and inference at 48, 64, and 80 pixels on AnyText-benchmark.}
\label{tab:atlas_full}
{\setlength{\tabcolsep}{3pt}
\begin{tabular}{llrrrrrrrr}
\toprule
 &  & \multicolumn{4}{c}{English} & \multicolumn{4}{c}{Chinese} \\
\cmidrule(lr){3-6}\cmidrule(lr){7-10}
Train px & Infer px & ACC $\uparrow$ & NED $\uparrow$ & FID $\downarrow$ & LPIPS $\downarrow$ & ACC $\uparrow$ & NED $\uparrow$ & FID $\downarrow$ & LPIPS $\downarrow$ \\
\midrule
48 & 48 & 85.08 & 0.9424 & 12.324 & 0.08367 & 75.53 & 0.8720 & 12.812 & 0.06769 \\
48 & 64 & 84.69 & 0.9397 & 12.474 & 0.08362 & 73.90 & 0.8662 & 12.830 & 0.06802 \\
48 & 80 & 84.29 & 0.9380 & 12.454 & 0.08390 & 69.69 & 0.8521 & 12.870 & 0.06875 \\
64 & 48 & 85.60 & 0.9409 & 12.503 & 0.08303 & 71.85 & 0.8592 & 13.234 & 0.06797 \\
64 & 64 & 85.89 & 0.9430 & 12.379 & 0.08289 & 76.00 & 0.8734 & 13.347 & 0.06782 \\
64 & 80 & 86.25 & 0.9433 & 12.476 & 0.08297 & 75.72 & 0.8718 & 13.339 & 0.06859 \\
80 & 48 & 84.60 & 0.9384 & 12.263 & 0.08290 & 60.95 & 0.8181 & 13.145 & 0.07025 \\
80 & 64 & 84.79 & 0.9402 & 12.297 & 0.08272 & 73.61 & 0.8650 & 13.312 & 0.06876 \\
80 & 80 & 85.44 & 0.9419 & 12.257 & 0.08284 & 75.72 & 0.8747 & 13.407 & 0.06818 \\
\bottomrule\end{tabular}}
\par\smallskip
\begin{minipage}{\linewidth}\footnotesize Human-final inputs and full-transcript scoring; original reference images, background compositing, lossless PNG. Guidance 1, 10 steps, seed 42; BF16 VAE.\end{minipage}
\end{table*}

\subsection{Training Objective}
\label{sec:supp_loss}
Tab.~\ref{tab:loss} ablates the text supervision term of Eq.~(7) of the main paper.
All rows use the mask-weighted flow matching loss $\mathcal L_{\rm flow}$ and differ in the pixel-space text supervision added to it: none, the feature-based perceptual loss of FLUX-Text~\cite{fluxtext} built on ODM~\cite{odm}, or our recognizer-based $\mathcal L_{\rm OCR}$, at a fixed weight or with the timestep weight $2t$.
These results demonstrate that:
1)~pixel-space text supervision remains indispensable alongside the atlas, as removing it drops ACC to 70.08 and 62.19, 15.8 and 13.8 points below \method-Bi;
2)~the recognizer-based $\mathcal L_{\rm OCR}$ is the more effective supervision, as at a fixed weight of 0.25 it reaches 84.79 and 75.62 against the best ODM setting at 83.30 and 70.27, a gain of 1.5 and 5.4 points, largest in Chinese, whose characters differ in exactly the stroke-level detail that recognizer features separate;
3)~a larger fixed weight trades image quality for accuracy, as raising $w$ from 0.25 to 1.0 adds at most 1.4 ACC points in English and 1.0 in Chinese while FID rises by 1.0 and 1.7;
4)~the timestep weight $2t$ obtains this accuracy at a lower cost in image quality, as $0.25\cdot 2t$ improves on the fixed 0.25 by 1.1 and 0.4 ACC points at nearly unchanged FID and LPIPS, and comes within 0.3 and 0.6 points of the most accurate fixed weights at an FID 0.9 and 1.5 lower than that of $w=1.0$, confirming the premise of Sec.~3.3 of the main paper that glyph structure is decided at high-noise timesteps.
We therefore set $\alpha=0.25$ with the timestep weight.
The FID and LPIPS that $\mathcal L_{\rm OCR}$ costs relative to ODM is the trade-off behind the LPIPS gap to FLUX-Text, which uses the ODM loss, in Tab.~1 of the main paper.

\section{Full Atlas Budget Results}
\label{sec:supp_budget}

A character cell costs tokens in proportion to its area: 9, 16, and 25 tokens at heights 48, 64, and 80 pixels, so the atlas grows linearly with the number of text units and with cell resolution.
This section specifies the token allocator used at inference, reports the full train--test grid of cell heights, gives the complete results behind the atlas budget paragraph in Sec.~4.3 of the main paper, and records the inference cost.

\paragraph{Token allocator.}
The allocator is designed for efficient inference: it fits the atlas to a target budget of $N$ tokens and, within that budget, raises the resolution of the cells that need clear stroke detail as far as possible.
A cell needs detail most when its glyph is small in the glyph image, which then under-resolves its strokes, and least when the glyph is already large there.
The allocator therefore ranks the cells by the size of their glyph in the glyph image, measured as the short side of its rendered strokes, and proceeds in three passes.
First, every cell starts at 64 pixels, and any cell whose glyph is already at least 80 pixels in the glyph image is lowered to 48, as the atlas adds little where the glyph image is well resolved.
Second, if the total still exceeds $N$, cells are lowered from 64 to 48 pixels in order of decreasing glyph size until it fits.
Third, the remaining budget promotes the smallest glyphs first, from 48 to 80 and then to 64 pixels, as far as $N$ allows.
Punctuation stays at 64 pixels throughout.
Every text unit is kept, so the budget is soft: under a 512-token budget, two of the 1{,}000 Chinese images still exceed it, the longest at 794 tokens.

\paragraph{Cell resolution.}
Tab.~\ref{tab:atlas} trains at five cell heights and tests each model at three.
These results demonstrate that:
1)~accuracy is insensitive to the cell height itself, as the three matched settings from 48 to 80 pixels lie within 0.8 ACC points in English (85.08, 85.89, 85.44) and 0.5 in Chinese (75.53, 76.00, 75.72), so a 48-pixel atlas, 44\% smaller than a 64-pixel one, costs under one point;
2)~English is insensitive to a train--test mismatch as well, staying within 82.77--86.25 over the whole grid, since Latin strokes remain several latent pixels wide even at 48 pixels;
3)~Chinese is sensitive to a mismatch, and the drop grows with the gap and is steepest when testing below the training height, from 75.72 to 60.95 for a model trained at 80 pixels and tested at 48, and from 75.33 to 48.18 for one trained at 96, as the strokes of complex characters no longer appear at the density the model learned to read them.
Cell resolution can therefore be chosen by budget, provided the model has seen the chosen heights during training, which is why the random resolution augmentation of Sec.~3.3 of the main paper draws from 48, 64, and 80 pixels, the range over which matched accuracy is flat.

\paragraph{Budget at inference.}
Tab.~\ref{tab:supp_budget} extends the atlas budget table of the main paper with 80-pixel cells, NED, FID, and LPIPS, and Tab.~\ref{tab:budget_length} gives the atlas length under each setting.
Both models start from the same checkpoint of \method\ after four epochs and are each trained for a fifth epoch with identical data order and noise, one with cells fixed at 64 pixels and one with random resolution augmentation.
They are evaluated with identical inputs and seeds.
These results demonstrate that:
1)~augmentation does not reduce accuracy at the default height, as the two models are within 0.2 ACC points at 64 pixels in both languages, at a cost of 0.28 and 0.40 in FID;
2)~English is insensitive to the budget, staying within 0.6 points of 64 pixels for both models under every setting;
3)~in Chinese, augmentation reduces the accuracy loss at 48 pixels from 5.21 to 3.11 points, at 80 pixels from 1.58 to 0.96, and under the allocator from 2.29 to 1.05, so it mitigates, without removing, the mismatch sensitivity of Tab.~\ref{tab:atlas}.
Under the allocator the longest atlas shrinks from 784 to 512 tokens in English and from 1{,}200 to 794 in Chinese, while the mean changes by under 11\% because the allocator also raises the cells of short atlases: its role is to bound the sequence length of text-heavy scenes, not to cut the average cost.

\paragraph{Inference cost.}
The atlas appends tokens to the two scene-sized streams, which already carry 1{,}024 tokens each, so the mean atlas length of 291 English and 175 Chinese tokens at 64 pixels is a small fraction of the sequence.
Measured on A800-80GB GPUs at $512\times512$ with 10 Euler steps, generation takes on average 5.18s per English and 4.30s per Chinese image at 64 pixels, with a per-process peak of about 32.1 GiB, dominated by the resident model.
Lowering to 48 pixels cuts the atlas by 44\% and shortens generation by about 3\%, and the allocator controls the long tail without a measurable change in mean runtime.
The extra cost over glyph-image-only conditioning is thus the appended atlas tokens, bounded by the allocator and traded against stroke detail through cell resolution.

\section{More Qualitative Results}
\label{sec:supp_qualitative}

\begin{figure*}[t]\centering
\includegraphics[width=.91\textwidth]{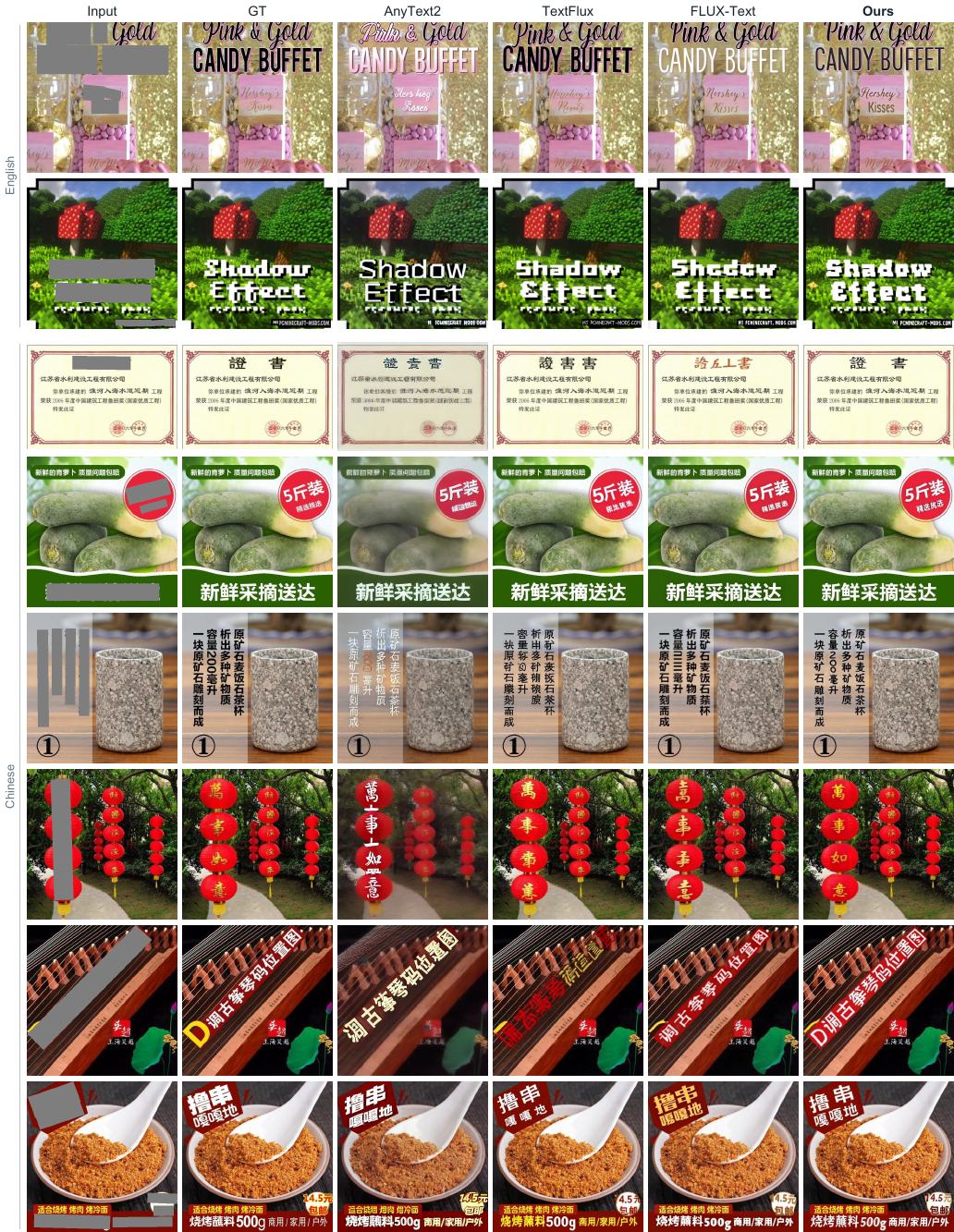}
\caption{\textbf{Scene text generation on AnyText-benchmark.}
Gray marks the regions to be filled, and ``GT'' abbreviates Ground Truth.}
\label{fig:supp_qual_gen}
\end{figure*}

\begin{figure*}[t]\centering
\includegraphics[width=.95\textwidth]{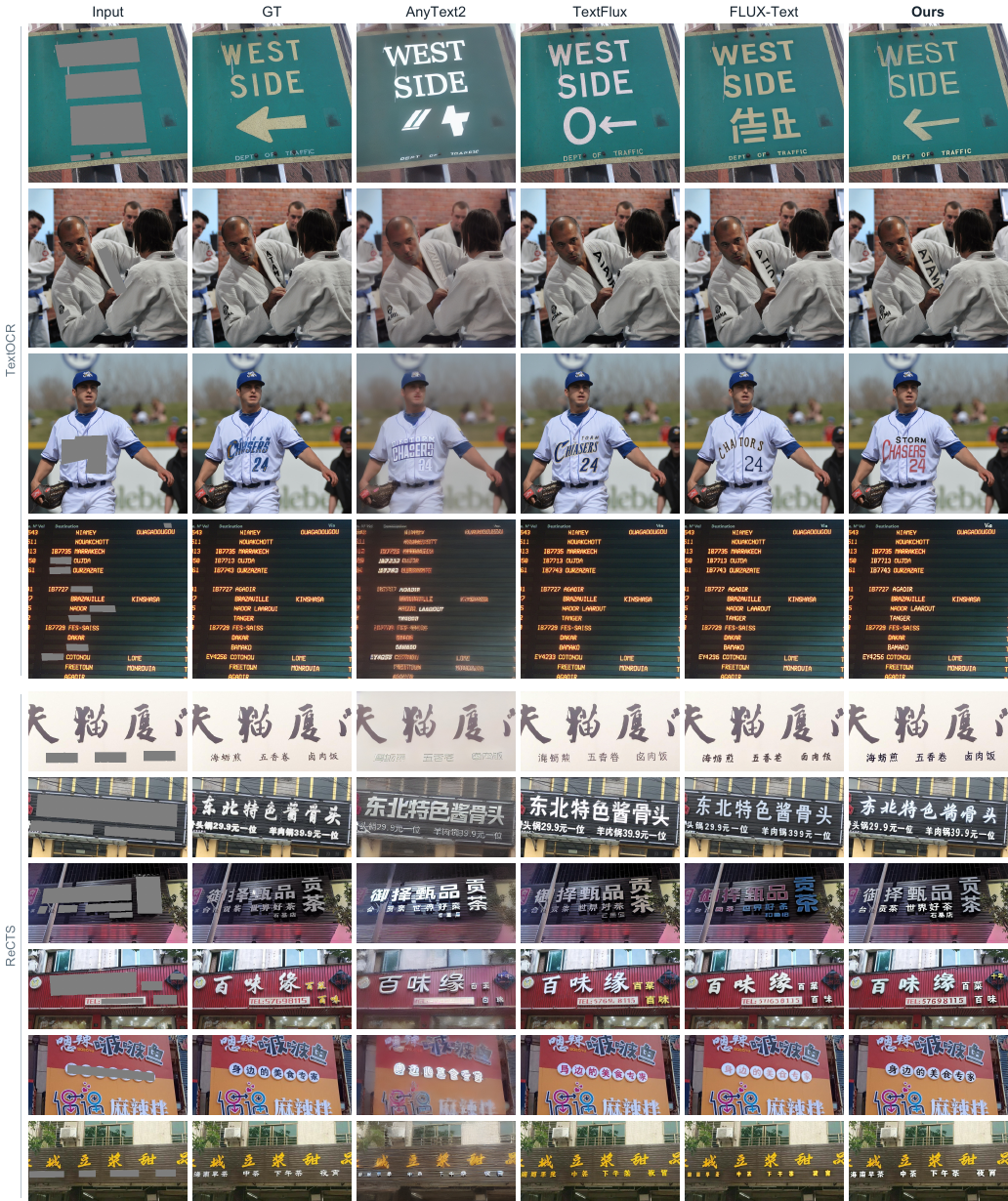}
\caption{\textbf{Scene text editing on TextOCR and ReCTS.}
Gray marks the regions to be filled, and ``GT'' abbreviates Ground Truth.}
\label{fig:supp_qual_edit}
\end{figure*}

We analyze qualitative results at three levels, moving from the benchmarks of the main paper to photographs beyond them: scene text generation (Fig.~\ref{fig:supp_qual_gen}), scene text editing (Fig.~\ref{fig:supp_qual_edit}), and out-of-distribution case studies (Fig.~\ref{fig:supp_gallery}).
Figs.~\ref{fig:supp_qual_gen} and~\ref{fig:supp_qual_edit} extend Fig.~4 of the main paper with the same six columns.
Every tile shows the complete image; only empty margins and the black letterbox bars of ReCTS are removed, at the same rows in every column.
Because the atlas cells share one uniform font, the remaining errors of \method\ in these cases concern font style, weight, or color rather than spelling.

\subsection{Scene Text Generation}
\label{sec:supp_qual_gen}
Fig.~\ref{fig:supp_qual_gen} shows two English and six Chinese images from AnyText-benchmark.
Chinese is where the glyph image is the weakest stroke reference, as each character must fit many strokes into the pixels its region allows, and it is where the baselines fail most.
On the lanterns, FLUX-Text repeats the second character of the four-character greeting and TextFlux replaces the third.
On the radish advertisement, both rewrite the four-character slogan, FLUX-Text in one character and TextFlux in all four.
On the vertical product description, AnyText2 turns the capacity of 200 into 600 and FLUX-Text replaces the digits with horizontal strokes, and on the zither chart, AnyText2 and FLUX-Text drop the leading letter D of the title.
On the seasoning advertisement, TextFlux drops the 1 from the price of 14.5 and FLUX-Text miswrites the most complex character of the product name.
The certificate shows a second weakness of the glyph image: its two characters are spread across a wide region, and the baselines fill the gap with a third or fourth character, whereas \method\ writes exactly two, consistent with the atlas holding exactly one cell per character.
\method\ renders every Chinese target correctly, in line with the gain of 8.1 ACC points on Chinese that the atlas brings over the glyph-image-only model in Tab.~4 of the main paper.
The English images test small and stylized Latin text: all three baselines misspell the cursive label on the candy jar, and on the game graphic FLUX-Text and TextFlux corrupt the letters of the pixel-art title and AnyText2 misspells the URL, whereas \method\ spells both correctly and keeps the pixel style of the title.

\subsection{Scene Text Editing}
\label{sec:supp_qual_edit}
Fig.~\ref{fig:supp_qual_edit} moves to real photographs from TextOCR and ReCTS, whose regions are small, numerous, and printed on varied surfaces.
Here a single wrong stroke often changes what the text means, and the baselines make exactly such errors.
FLUX-Text drops the decimal point of a price, turning 39.9 into 399, TextFlux and FLUX-Text write EY4233 and EY4236 for the flight EY4256, and all three baselines garble, drop, or alter digits of the eight-digit phone number.
Symbols fail in the same way: on the street sign, AnyText2 turns the arrow into an unrelated white shape, TextFlux adds a letter O before it, and FLUX-Text replaces it with two glyphs that resemble Chinese characters.
On the six ReCTS signs, the baselines miswrite characters from the three dish names of the snack bar to the slogan on seven round badges and the small branch name of the tea shop, all of which \method\ renders legibly.
Chinese text in small regions is the condition behind the largest margin of \method\ in Tab.~2 of the main paper, on ReCTS, and several signs here carry five or six regions, the cluttered setting in which atlas-anchored attention outperforms full attention by 4.0 points in Tab.~4.
On clothing, \method\ also spells the words on the martial-arts uniform and the baseball jersey correctly, where FLUX-Text merges the two words of the jersey and TextFlux writes STORN.

\subsection{Out-of-Distribution Case Studies}
\label{sec:supp_qual_ood}
\begin{figure*}[p]\centering
\includegraphics[width=\textwidth]{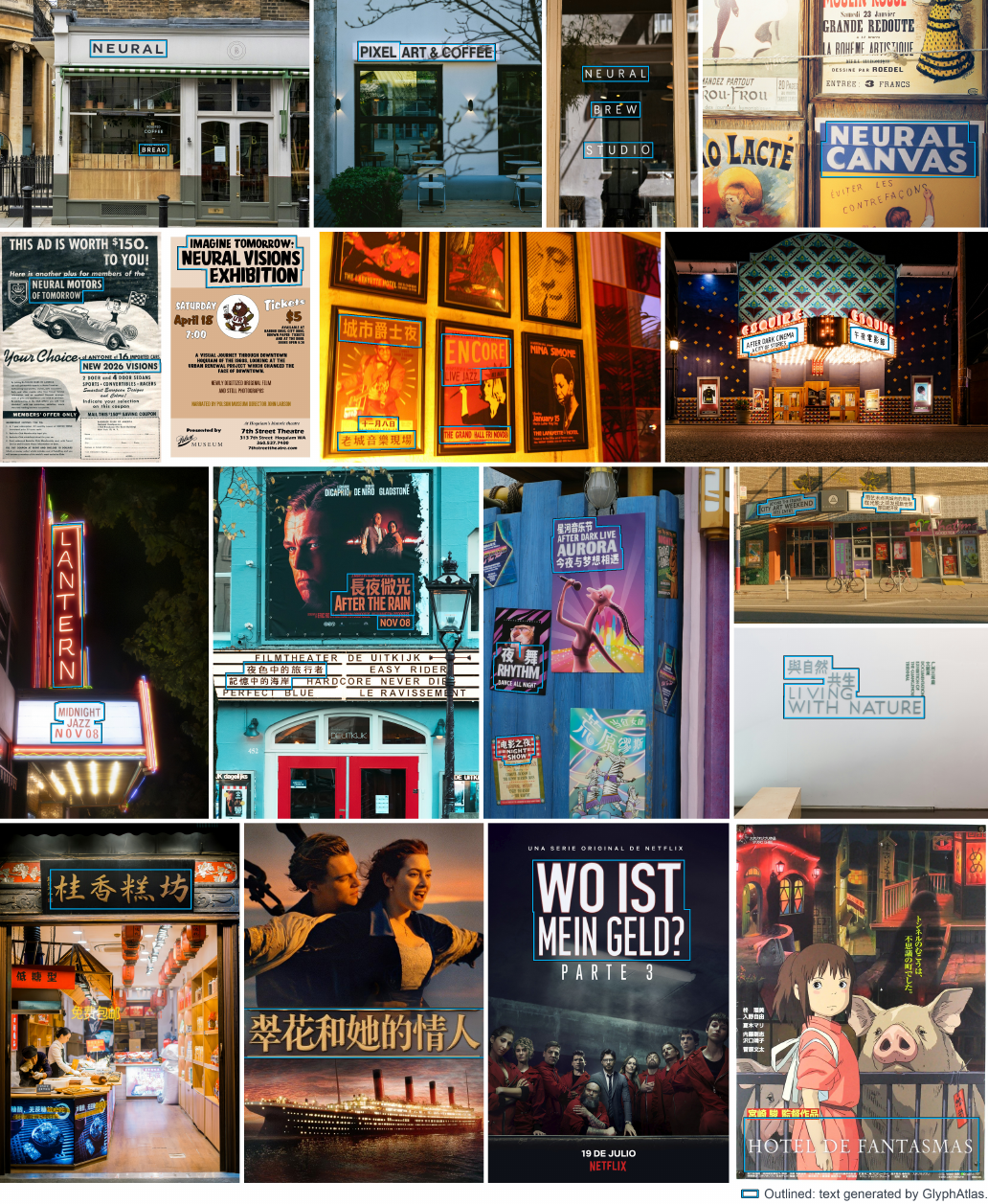}
\caption{\textbf{Further case studies.}
Outlines mark the text generated by \method.}
\label{fig:supp_gallery}
\end{figure*}

Fig.~\ref{fig:supp_gallery} leaves the benchmarks and shows further \method\ case studies that do not appear in Fig.~1 of the main paper.
Each case is a single editing pass on a photograph, a scanned print, or a released poster at its native resolution, from 768 to 1{,}920 pixels per side, whereas training uses $512\times512$ crops.
As in Fig.~1, every tile is the complete output and the outlines mark the generated regions.
The cases cover what the benchmarks do not: storefronts, shop windows and painted signs, whose text must take on the material and lighting of its carrier; vintage print advertisements and event posters, whose display typefaces are far from the uniform font of the atlas; bilingual marquees and poster walls with up to ten fields; and a calligraphic shop sign and film posters retitled in another language.
The generated text follows its carrier in each case, from the raised letters of the caf\'e sign and the single tubes of the neon sign to the printed ink of the vintage advertisement and the gilded strokes of the calligraphic sign.
Dense Chinese text stays legible at poster scale, including the traditional characters of the concert poster and the ten Chinese and English fields of the three night-event posters, all written in one pass.
On the retitled film posters, the new Chinese, German, and Spanish titles take the color and weight of the poster design, and the Japanese title of the last poster is left untouched.

\end{document}